\documentclass[10pt,twocolumn,letterpaper]{article}

\usepackage[pagenumbers]{wacv} 

\usepackage{graphicx}
\usepackage{amsmath}
\usepackage{cite}
\usepackage{booktabs}
\usepackage{multirow}
\usepackage{makecell}
\usepackage{pifont}
\usepackage{dsfont}
\usepackage{amsmath, amssymb}
\usepackage{xcolor}
\usepackage[export]{adjustbox}
\usepackage{hhline}

\newcolumntype{P}[1]{>{\centering\arraybackslash}p{#1}}
\newcolumntype{L}[1]{>{\leftalign\arraybackslash}p{#1}}

\usepackage[pagebackref,breaklinks,colorlinks]{hyperref}

\usepackage[capitalize]{cleveref}
\crefname{section}{Sec.}{Secs.}
\Crefname{section}{Section}{Sections}
\Crefname{table}{Table}{Tables}
\crefname{table}{Tab.}{Tabs.}

\def\wacvPaperID{308} 
\def\confName{WACV}
\def\confYear{2025}

\begin{document}

\title{Feature Attenuation of Defective Representation \\Can Resolve Incomplete Masking on Anomaly Detection}

\author{YeongHyeon Park$^{1, 2}$ \qquad Sungho Kang$^{1}$ \qquad Myung Jin Kim$^{2}$ \\
\qquad Hyeong Seok Kim$^{2}$ \qquad Juneho Yi$^{1}$~\thanks{Corresponding author: Juneho Yi.} \\
$^{1}$Department of Electrical and Computer Engineering, Sungkyunkwan University\\
$^{2}$SK Planet Co., Ltd.\\
{\tt\small \{yeonghyeon, myungjin, beman\}@sk.com, \{sungho369, jhyi\}@skku.edu }
}
\maketitle

\begin{abstract}
In unsupervised anomaly detection (UAD) research, while state-of-the-art models have reached a saturation point with extensive studies on public benchmark datasets, they adopt large-scale tailor-made neural networks (NN) for detection performance or pursued unified models for various tasks. 
Towards edge computing, it is necessary to develop a computationally efficient and scalable solution that avoids large-scale complex NNs. Motivated by this, we aim to optimize the UAD performance with minimal changes to NN settings. Thus, we revisit the reconstruction-by-inpainting approach and rethink to improve it by analyzing strengths and weaknesses. 
The strength of the SOTA methods is a single deterministic masking approach that addresses the challenges of random multiple masking that is inference latency and output inconsistency. Nevertheless, the issue of failure to provide a mask to completely cover anomalous regions is a remaining weakness. 
To mitigate this issue, we propose \textbf{F}eature \textbf{A}ttenuation of \textbf{De}fective \textbf{R}epresentation (\textbf{FADeR}) that only employs two MLP layers which attenuates feature information of anomaly reconstruction during decoding. By leveraging FADeR, features of unseen anomaly patterns are reconstructed into seen normal patterns, reducing false alarms.
Experimental results demonstrate that FADeR achieves enhanced performance compared to similar-scale NNs. Furthermore, our approach exhibits scalability in performance enhancement when integrated with other single deterministic masking methods in a plug-and-play manner.
\end{abstract}

\vspace{-0.5cm}
\section{Introduction}

\begin{figure}[t]
    \scriptsize
    \setlength{\tabcolsep}{0pt}
    \centering
    \resizebox{0.95\linewidth}{!}{%
        \begin{tabular}{c}
            \begin{tabular}{c ccc}
                \quad{} & \multirow{2}{*}{Input} & Binary mask & Soft mask \\
                \quad{} & & of pre-trained attention & of FADeR \\
                
                \rotatebox[origin=c]{90}{\qquad{}\qquad{}\qquad{}\qquad{}\quad{}Cable} &
                \includegraphics*[width=0.278\columnwidth]{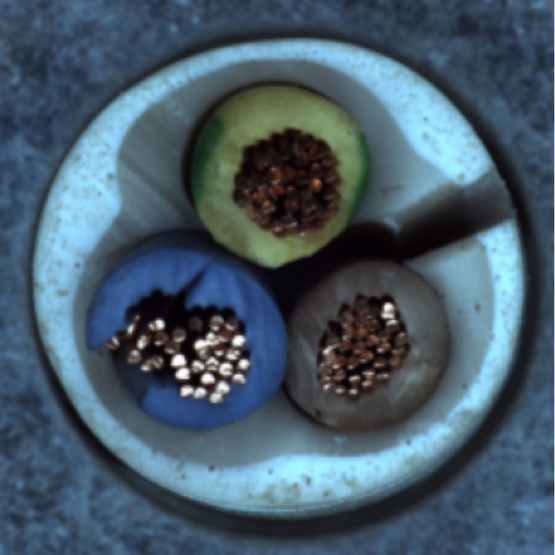} &
                \includegraphics*[width=0.278\columnwidth]{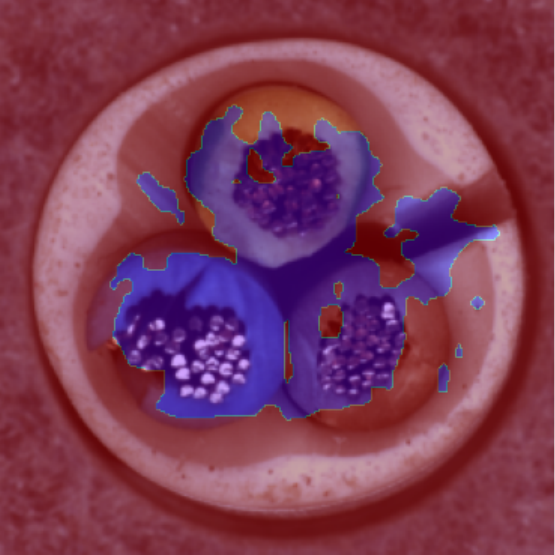} &
                \includegraphics*[width=0.278\columnwidth]{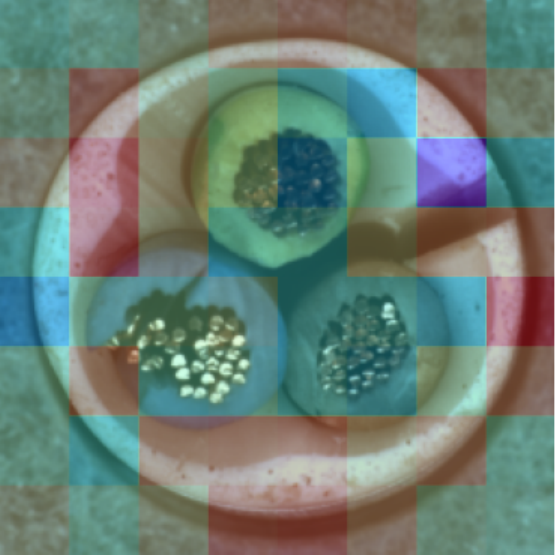} \\
                \vspace{-1.55cm} \\
                
                \rotatebox[origin=c]{90}{\qquad{}\qquad{}\qquad{}\qquad{}\quad{}Transistor} &
                \includegraphics*[width=0.278\columnwidth]{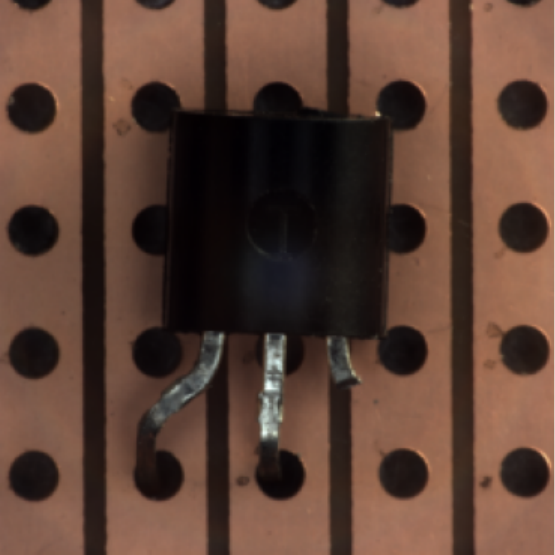} &
                \includegraphics*[width=0.278\columnwidth]{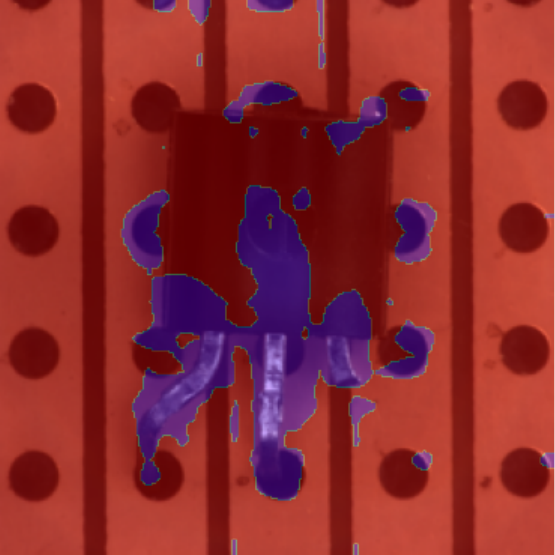} &
                \includegraphics*[width=0.278\columnwidth]{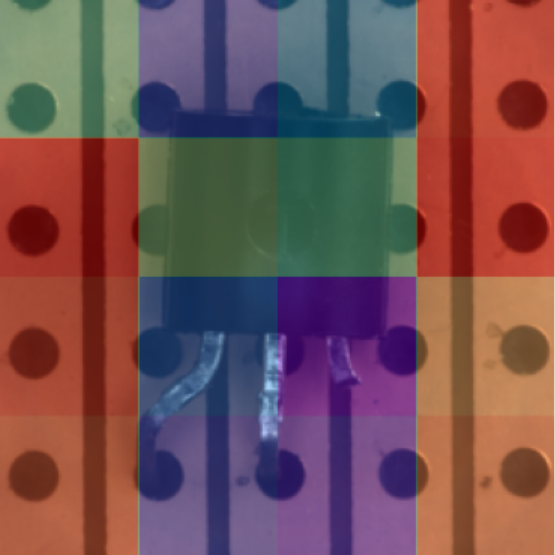} \\
                \vspace{-1.75cm} \\

                \rotatebox[origin=c]{90}{\qquad{}\qquad{}\qquad{}\qquad{}\qquad{}Capsule} &
                \includegraphics*[width=0.278\columnwidth]{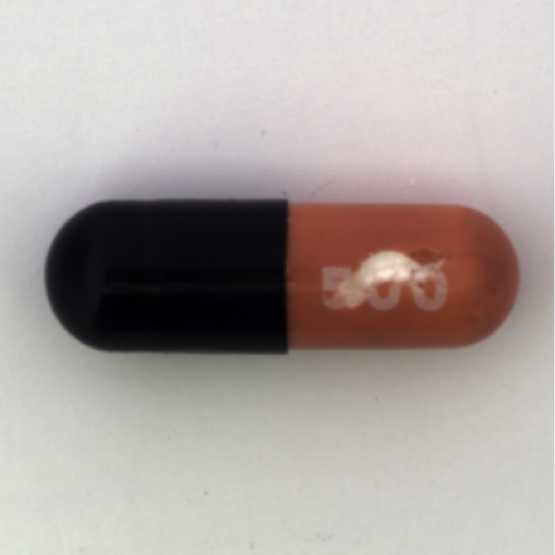} &
                \includegraphics*[width=0.278\columnwidth]{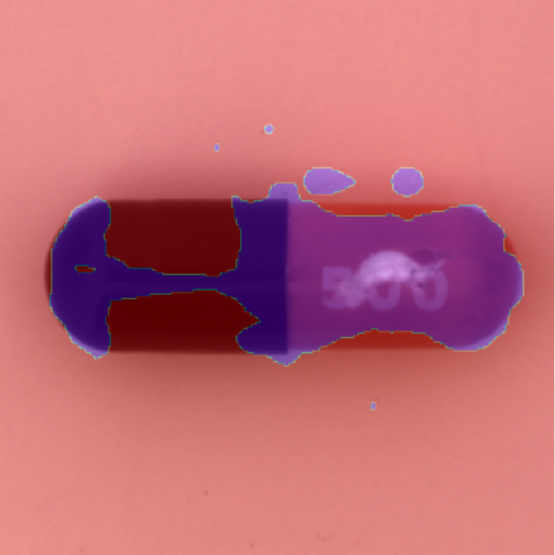} &
                \includegraphics*[width=0.278\columnwidth]{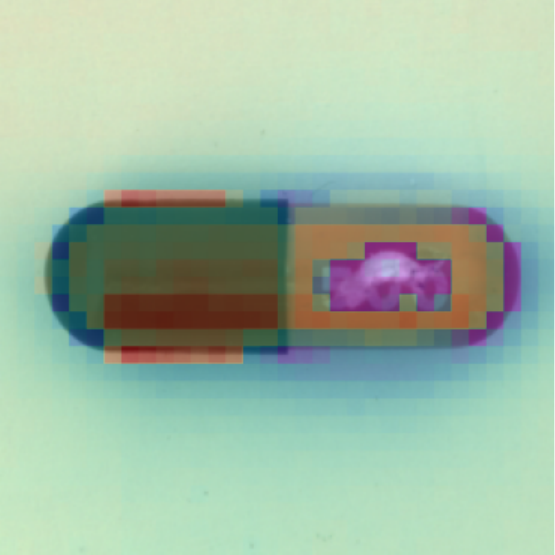} \\
                \vspace{-1.67cm} \\
            \end{tabular}
            \\
            \includegraphics*[width=0.8\columnwidth,trim={1cm 1.5cm 1.3cm 9.3cm},clip]{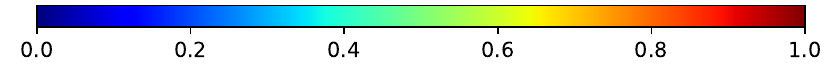} \\
            \vspace{-0.6cm} \\
        \end{tabular}
    }
    \caption{Binary masks of pre-trained attention and soft masks of FADeR. Masks are applied to images or feature maps as an element-wise product. The color bar represents the value of the mask. When the value of the mask is closer to 0 (blue color), the attenuation is stronger. The soft mask attenuates feature representations of missing defective part of the incomplete binary attention mask. Although the binary mask is larger than the defective regions as shown in the capsule case, FADeR further will help cut out the core part of the defective representation on skip connections.}
    \label{fig:vis_results}
    \vspace{-0.5cm}
\end{figure}

\begin{figure}[t]
    \centering
    \resizebox{\linewidth}{!}{%
        \includegraphics*[width=1.0\linewidth,trim={0.0cm 0.0cm 0.0cm 0.0cm},clip]{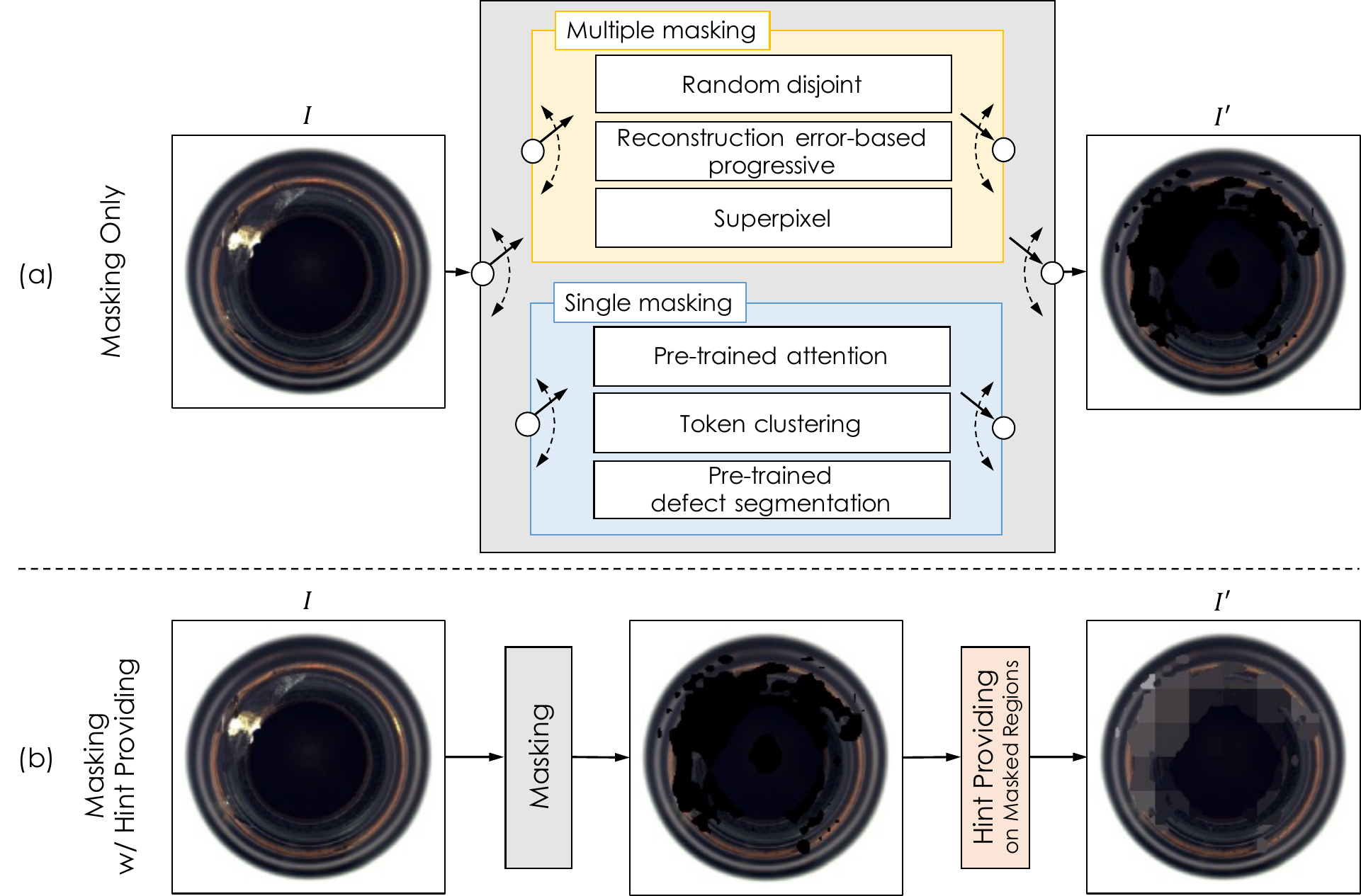}
    }
    \vspace{-0.25cm}
    \caption{Visual defect masking methods in reconstruction-by-inpainting for UAD. (a) shows a method for masking suspected defects. We can optionally adopt a masking method. We adopt a single masking method because multiple masking methods suffer in inference latency. (b) shows a method of providing visual obfuscation-based hints in the masked regions. We leverage a strategy shown in (b) because it further improves the UAD performance of (a) by deriving accurate normal reconstruction~\cite{EAR_Park_arXiv23}.}
    \label{fig:obfuscation}
    \vspace{-0.3cm}
\end{figure}

When building an automated anomaly detection (AD) system, we first encounter a significant challenge of data imbalance problem due to the rarity of anomaly situations in the early stages of manufacturing.
This has been dealt with by adopting an unsupervised anomaly detection (UAD) method that only exploits prevalent normal samples for the training.
In recent years, UAD researches have shown significant advancements, particularly with the exploration of state-of-the-art (SOTA) models on public benchmark datasets~\cite{RIAD_Vitzan_PR21,DSR_ZavrtanikK_ECCV22,OmniAL_Zhao_CVPR23,EAR_Park_arXiv23,AMINet_Luo_TASC24,AMINet_Luo_TASC24,CRAD_Lee_arXiv24}. 
The focus has often revolved around the use of large-scale tailored neural networks (NN) to achieve better AD performance~\cite{GANomaly_Akcay_ACCV18,MemAE_Gong_ICCV19,BWMem_Hou_ICCV21,RD4AD_Deng_CVPR22,APMask_Bozorgtabar_AAAI23} or the development of unified models~\cite{UniAD_You_Neurips22,HVQ-Trans_Lu_Neurips23} capable of handling multiple tasks that are trained with known multiple-category products. 
However, towards edge computing environments, there is a growing need to shift towards methods that are not only computationally efficient but also scalable.

One key observation from current SOTA methods is their reliance on single deterministic masking methods~\cite{EAR_Park_arXiv23,AMINet_Luo_TASC24} to address challenges related to random or multiple masking methods~\cite{RIAD_Vitzan_PR21,BWMem_Hou_ICCV21,Randommask_Nardin_ICIAPW22,MAEMRI_Lang_arXiv23,SSM_Huang_Tran.Multimedia23}, which often result in increased inference latency and output inconsistency. 
However, incomplete mask issues can occur while leveraging a single deterministic masking method.
First of all, collection of large quantities of abnormal samples is infeasible and cannot cover all types of defects that may occur at test time.
Therefore, creating a model that can perfectly mask defective regions is practically impossible.
For another example, consider a pre-trained attention-based zero-shot masking method~\cite{EAR_Park_arXiv23}.
It presents an attractive feature for masking suspected defective regions but is partially imperfect to mask defects.
This is because it was not trained to mask defects.
Therefore, we need to warrant that any single deterministic masking method can fully cover defective regions.

To address this issue, we introduce an effective method to resolve the possible missing out of defective regions when exploiting single deterministic masking in the reconstruction-by-inpainting-based UAD approach.
Our study aims to optimize UAD performance with minimal changes to the NN structure by using a reconstruction-by-inpainting U-Net, which effectively prevents the identity shortcut (IS) issue. 
We propose further refinements to enhance its efficacy through a detailed analysis.
Our method, dubbed \textit{\textbf{F}eature \textbf{A}ttenuation of \textbf{De}fective \textbf{R}epresentation} (FADeR), leverages a simple architecture comprising only two Multi-Layer Perceptron (MLP) layers to attenuate defective feature information by soft feature masking during the decoding process. 
FADeR divides a visual defect obfuscated image into patches and predicts the patch-wise reconstruction errors.
At this time, a high value of reconstruction error means that the patch is very likely to the defective region.
FADeR attenuates the encoded feature representation of suspected defective patches before transmitting those to the decoder.
The decoder receives a feature map with attenuated defective representation, which can promote reconstruction into a seen normal pattern.
Note that, we exploit the active learning strategy~\cite{AL_Yoo_CVPR19} to train FADeR because there is no label of defective feature attenuation.
Active learning allows training NNs with minimal or no labeled data.

Experimental results with the public industrial datasets MVTec AD~\cite{MVTec_Bergmann_CVPR19} and VisA~\cite{VisA_Zou_ECCV22} demonstrate the capability of FADeR that resolves the incomplete mask issue of single deterministic masking methods.
Furthermore, our approach suggests scalability and versatility by exhibiting improved performance when integrated with other UAD models in a plug-and-play manner.

Overall, our contributions are summarized as follows:
\vspace{-0.15cm}
\begin{itemize}
    \item We propose an effective solution to resolve the possible incomplete mask issue that may not fully cover defective regions. 
    This solution mitigates unintended false negatives caused by the features of unmasked defective regions during the masking process in the reconstruction-by-inpainting approach.
    Additionally, this component only works as a two-layer MLP, which can minimize increase in computational complexity.
    \vspace{-0.2cm}

    \item We propose FADeR as a scalable solution by allowing us to plug and play with other inpainting design components or freely adjust the detailed design of FADeR components.
\end{itemize}

\begin{figure}[t]
    \resizebox{\linewidth}{!}{%
        \includegraphics*[width=1.0\linewidth,trim={0.0cm 0.0cm 0.0cm 0.0cm},clip]{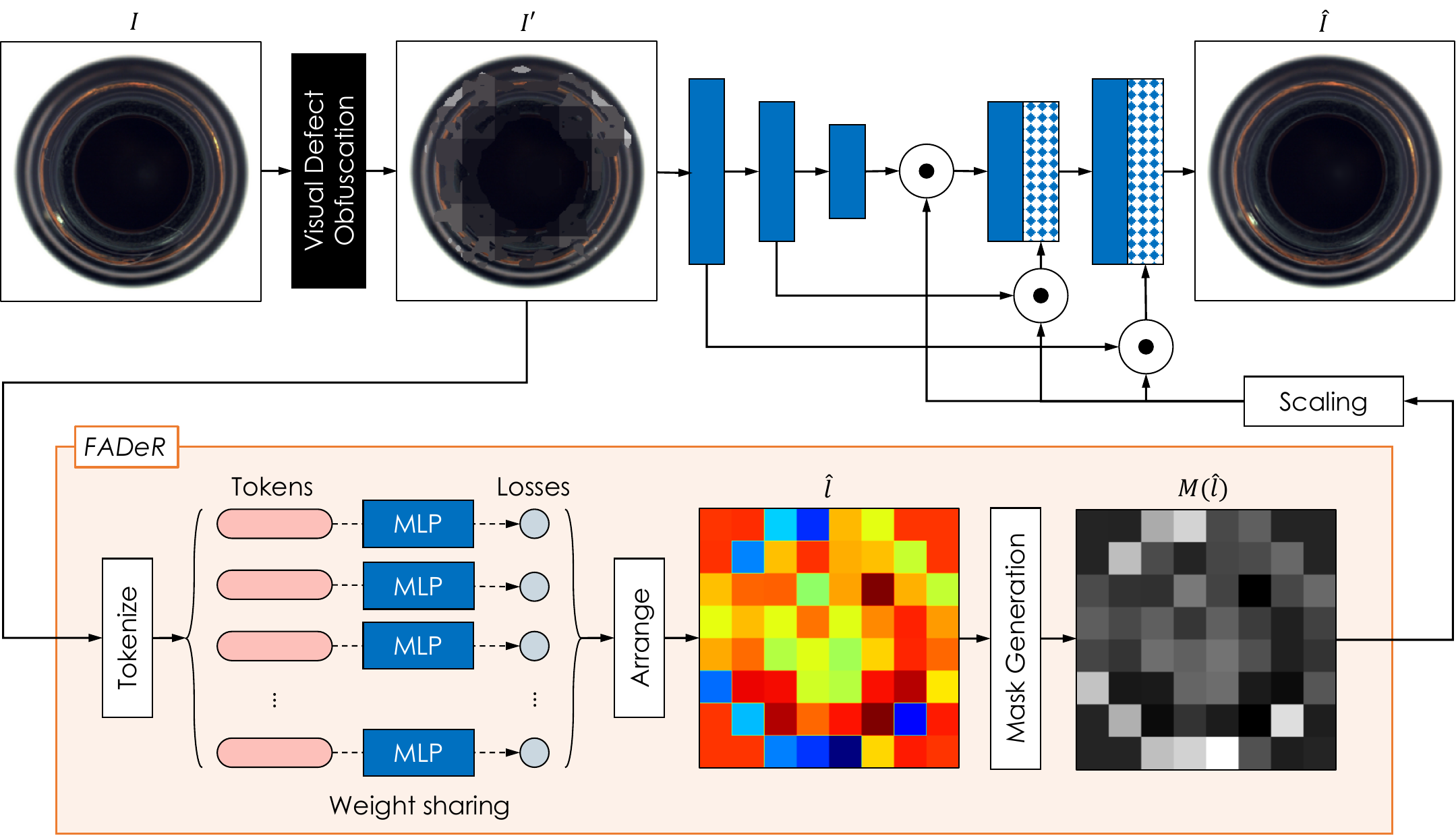}
    }
    \vspace{-0.3cm}
    \caption{The overview of our method, FADeR. We propose a simple two-layer MLP component to overcome the incomplete mask issue in single deterministic masking of visual defect obfuscation process. The visual obfuscation method is shown in Figure~\ref{fig:obfuscation}. It attenuates defective feature representations in skip connections. The training detail of FADeR is shown in Figure~\ref{fig:prep_loss}.}
    \label{fig:prep_training}
\end{figure}

\section{Related work}
The detection of anomaly instances, known as AD, has garnered significant interest because it is a critical task across a wide range of industries. Unsupervised anomaly detection is a widely adopted approach due to anomalous data scarcity and difficult labeling situations.

\subsection{Reconstruction-by-inpainting methods}
An UAD model can be trained by minimizing the reconstruction error with normal samples only. 
Then, we can find anomalous samples in the test stage by the reconstruction error between the input and output of the model.
Many recent studies dealt with the strategy that changes NN structure to improve the UAD performance.
U-Net like structures that integrates skip connections into the autoencoder (AE) are utilized in recent studies~\cite{ADSC_Collin_ICPR21,DaA_Hou_ICCV21,EdgRec_Liu_arxiv22}.
Other strategies to construct NNs include: 
1) replacing each layer within AEs with other types such as convolutions, MLPs, or transformers\cite{VT-ADL_Mishra_ISIE21,ADTR_You_Neurips23,AnoVit_Kang_Access22,IIAD-VT_Yang_sensors24,HaloAE_mathian_arxiv22,InTra_Pirnay_ICIAP22,MaskedSwin_Jiang_TII23,MaskedTransformer_Nardin_IJNS22,UniAD_You_Neurips22}, 
2) adding other branches of neural propagation such as generative adversarial networks (GAN)\cite{SCADN_Yan_AAAI21,OCR-GAN_Liang_TIP23}, and 
3) changing the computational flow without structural change of the NN for example normalizing flow or diffusion process\cite{AnoDDPM_Wyatt_CVPR22,DiffAD_Zhang_ICCV23,DiAD_He_AAAI24,MSFlow_Zhou_TNNLS24}.
These recent advancements are indeed adequate efforts to achieve high performance.
However, they require huge computational resources, which can be a drawback for practical applications in industrial settings. 

Reconstruction-by-inpainting methods have emerged to achieve better performance by effectively preventing an UAD model from accurate reconstruction of unseen anomalous patterns~\cite{RIAD_Vitzan_PR21,Randommask_Nardin_ICIAPW22,MaskedSwin_Jiang_TII23,MAEMRI_Lang_arXiv23,SSM_Huang_TM22,Iter_Hitoshi_JIP22,InTra_Pirnay_ICIAP22,MSR_Cosmin_ICMLW23,EAR_Park_arXiv23,AMINet_Luo_TASC24}.
These are visually summarized in Figure~\ref{fig:obfuscation}.
Specifically, they can be categorized into three cases, random masking~\cite{RIAD_Vitzan_PR21,Randommask_Nardin_ICIAPW22,MaskedSwin_Jiang_TII23,MAEMRI_Lang_arXiv23}, multiple disjoint masking~\cite{RIAD_Vitzan_PR21,SSM_Huang_TM22}, and progressive masking from the initial masks~\cite{SupPix_Li_BMVC20,Iter_Hitoshi_JIP22,SSM_Huang_TM22,InTra_Pirnay_ICIAP22,MSR_Cosmin_ICMLW23}.
While multiple random masking methods ensure a good performance, they commonly lead to inference latency due to multiple masking and output inconsistency due to random masking.

To address this issue, methods of single deterministic masking are developed.
EAR~\cite{EAR_Park_arXiv23} leverages the attention of ImageNet pre-trained DINO-VIT~\cite{DINO_Caron_ICCV21} to generate a single deterministic mask.
It has a strength that does not need a training process for mask generation.
On the other hand, AMI-Net~\cite{AMINet_Luo_TASC24} proposes a clustering-based learnable single masking method.
The clustering process of AMI-Net~\cite{AMINet_Luo_TASC24} allows spatial context-aware masking through token embedding.
These methods have a strength of output consistency and fast inference.
However, compared to multiple random masking methods, it can potentially fail to provide a mask that completely cover defective regions.

To resolve this problem we propose a design component that effectively complements previous binary masking that masks defective features in the decoding process. 
In addition, to meet the computational efficiency, we make it an extremely tiny design by employing MLP with two layers only.

\begin{figure}[t]
    \centering
    \resizebox{\linewidth}{!}{%
        \includegraphics*[width=1.0\linewidth,trim={0.0cm 0.0cm 0.0cm 0.0cm},clip]{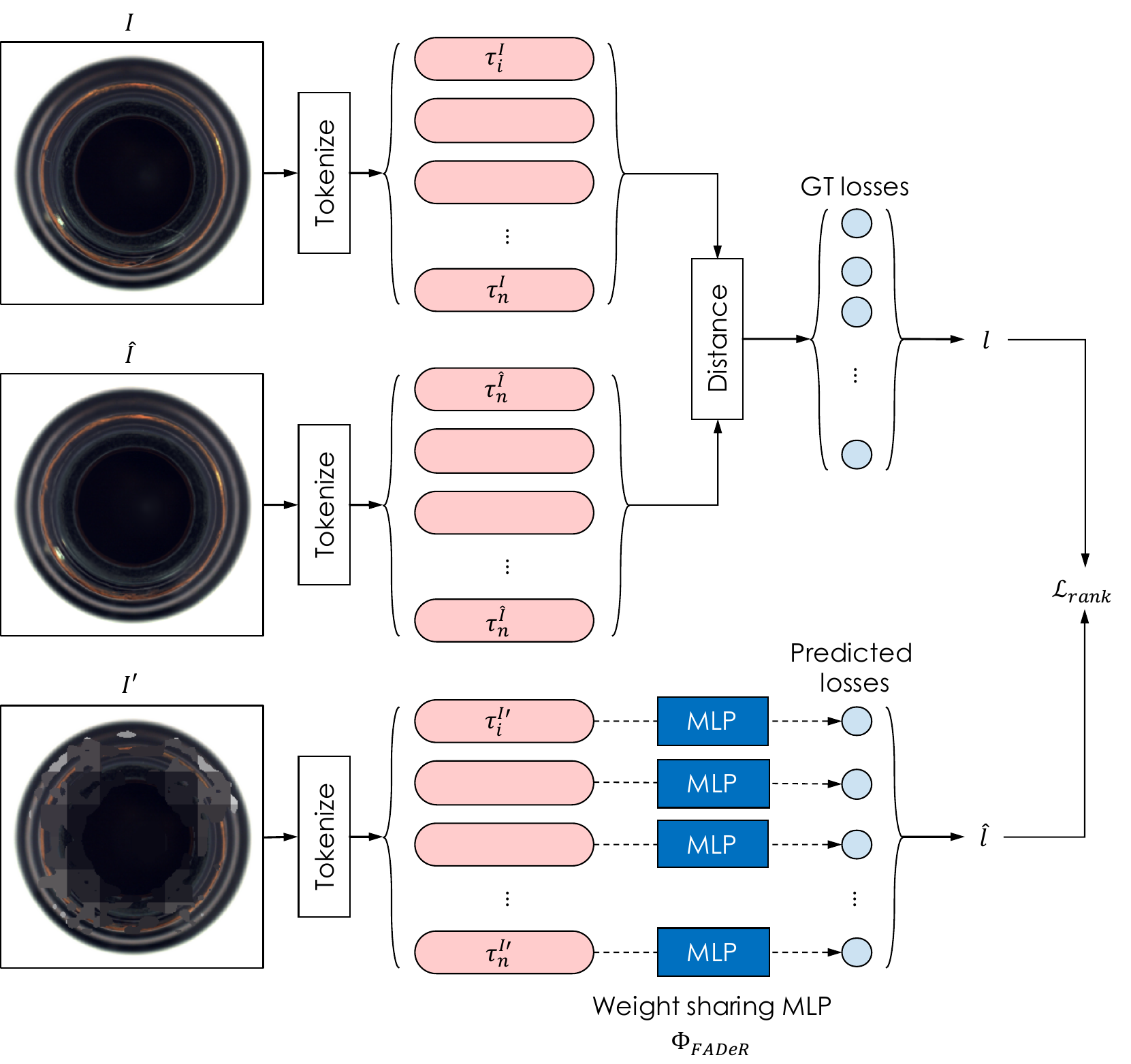}
    }
    \vspace{-0.3cm}
    \caption{Construction of ground truth (GT) loss for training FADeR with only normal samples. After generating GT loss using $I$ and $\hat{I}$, FADeR learns to predict GT loss from $I'$ via MLP with loss function as~\eqref{eq:loss_rank}.}
    \label{fig:prep_loss}
\end{figure}

\subsection{Feature representations on UAD}
While most UAD methods image-level reconstruction, some recent studies choose a feature-level reconstruction approach.
This method generates an anomaly score by measuring the distance between feature maps.
Their experimental results show that defective feature representations are effectively converted into normal feature representations to help reduce false detection~\cite{DSR_ZavrtanikK_ECCV22,CRAD_Lee_arXiv24}.
They also proposed a way to avoid an IS issue caused by skip connections.
DSR~\cite{DSR_ZavrtanikK_ECCV22} addresses the IS issue by leveraging a latent vector codebook of quantized normal representations for subspace reprojection.
HVQ-Trans~\cite{HVQ-Trans_Lu_Neurips23} point out the importance of vector quantization (VQ)~\cite{VQVAE_Aaron_NIPS17} in preventing the IS issue and propose a codebook switching framework to replenish frail normal patterns that can be caused by codebook collapse.

CRAD~\cite{CRAD_Lee_arXiv24} exploits continuous memory of the normal features to replace suspected abnormal features with the synthetic normal feature representations. 
They select suspected abnormal features using combined distance of cosine similarity and mean squared error between original input features and fused normal features that are synthesized from the continuous memory of normal.

We propose a strategy to attenuate defective representations by predicting suspected defective patches to address the IS issue.
At the same time, we make it an easy-to-access design, two-layered MLP, for plug-and-play with other existing models.

\section{Methods}

\subsection{Overview}
An overview of FADeR is shown in Figure~\ref{fig:prep_training}.
We integrate the proposed scalable cooperative NN component, dubbed FADeR, with a pre-trained UAD model to address the incomplete mask issue. 
FADeR generates a soft mask of suspected defective features from visually defective obfuscated image $I'$.
This soft mask is placed on skip connections to avoid an IS issue.
As a result, accurate reconstruction of anomalous patterns by incomplete mask will be mitigated by applying our method.

\begin{figure}[t]
    \centering
    \resizebox{\linewidth}{!}{%
        \includegraphics*[width=1.0\linewidth,trim={0.0cm 0.0cm 0.0cm 0.0cm},clip]{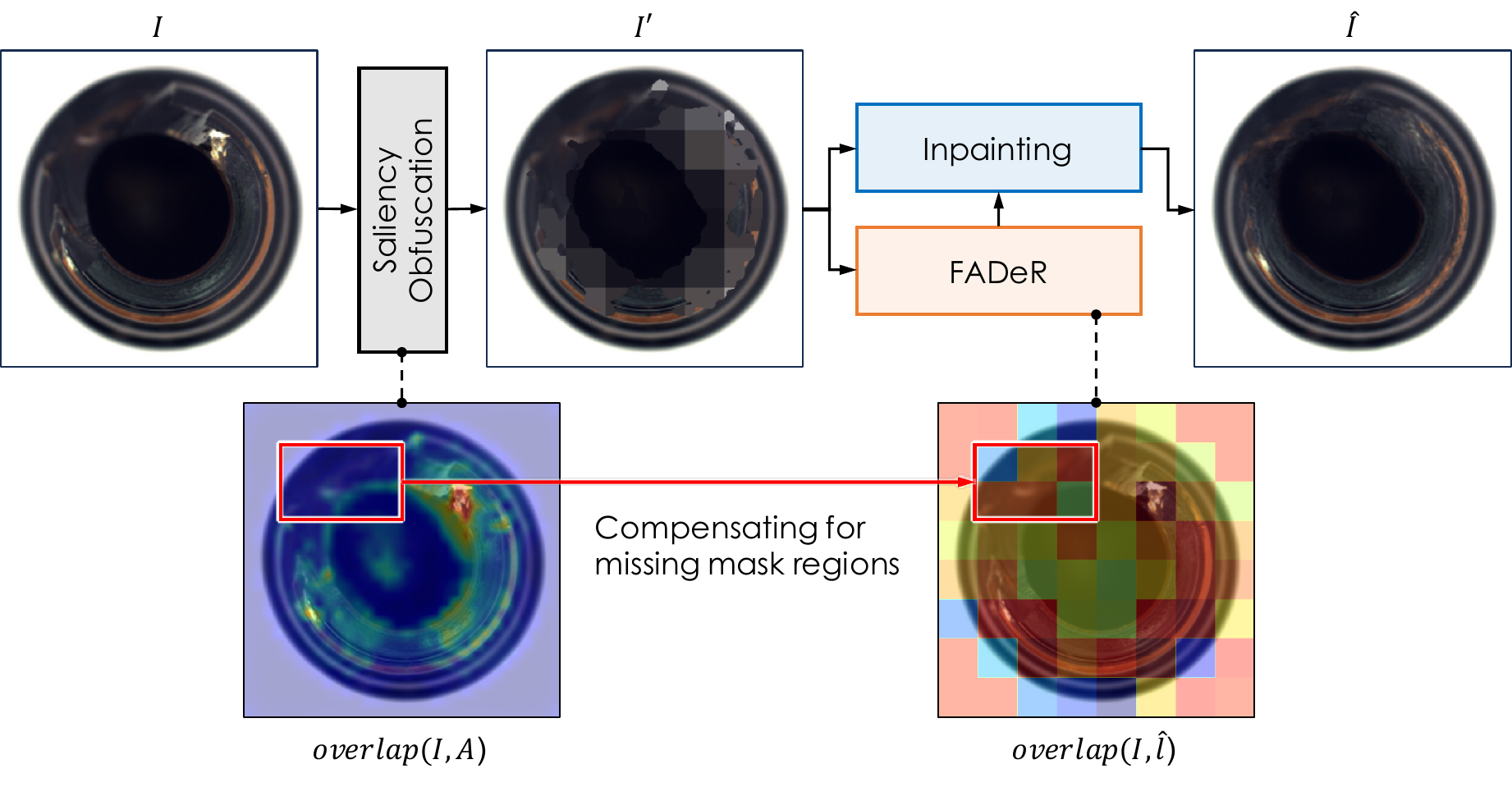}
    }
    \vspace{-0.2cm}
    \caption{Inference process of FADeR. FADeR attenuates visual defective features by predicting patch-wise error.}
    \label{fig:prep_inference}
\end{figure}

\subsection{Method to resolve incomplete mask}
Our method, FADeR, focuses on attenuating defective features possibly delivered through skip connection in incomplete mask issues.
At this time, the degree of attenuation varies depending on the predicted patch-wise error by FADeR. 
To train FADeR, we should provide ground truth (GT) of feature attenuation but there is no label since it is hard to interpret the feature-level information by humans.
Thus, we need to find a breakthrough to train FADeR without human label.

We take the active learning strategy that allows training NNs with minimal or no labeled data~\cite{AL_Yoo_CVPR19}. 
By leveraging this, the GT for FADeR training will be constructed by the reconstruction error between the input image $I$ and the reconstruction result $\hat{I}$ in a patch-wise manner.
Then, we update FADeR to predict the target GT and iterate these two steps during the target training epoch.
The graphical representation of this procedure is shown in Figure~\ref{fig:prep_loss}.

For patch-size operation, we need to set the patch size.
In (b) of Figure~\ref{fig:obfuscation}, the mosaic obfuscation-based hint is provided on masked regions to promote the accurate reconstruction of normal patterns.
Note that, an optimal optimal mosaic scale for each product is found and provided by EAR~\cite{EAR_Park_arXiv23}.
However, when the mosaic scale and patch size are different from each other, unintended hint of defective informations will be partially passed through skip connection and degrades UAD performance.
Thus, we set the patch size for tokenization equal to the mosaic scale to spatially synchronize patch token of FADeR and mosaic-based hint.

By referring to the above settings, the input $I$ and the reconstruction result $\hat{I}$ are divided into patch units and tokenized.
These are represented as $\tau{}^{I}_{i}$ $\tau{}^{\hat{I}}_{i}$ with an index $i$.
Sequentially, the $\mathcal{L}_{2}$ between $\tau{}^{I}_{i}$ $\tau{}^{\hat{I}}_{i}$ is calculated to generate the GT loss $l_{i}$.
The set of patch-wise GT loss is denoted by $l$ as ~\eqref{eq:loss_token}.
Patch-wise error prediction through FADeR also begins with the tokenization process.
The two-layer MLP, $\Phi{}_{\textit{FADeR}}$, performs loss prediction for each token $\tau{}^{I'}_{i}$ as ~\eqref{eq:pred_fader}.

\vspace{-0.3cm}
\begin{equation}
    \begin{aligned}
        l = \{l_{i},\ \  1\leq{}i\leq{}n\} \quad{} \text{s.t.}\quad{} l_{i} = \mathcal{L}_{2}(\tau{}^{I}_{i}, \tau{}^{\hat{I}}_{i}) 
    \end{aligned}
    \label{eq:loss_token}
\end{equation}

\vspace{-0.3cm}
\begin{equation}
    \begin{aligned}
        \hat{l} = \{\hat{l}_{i},\ \  1\leq{}i\leq{}n\} \quad{} \text{s.t.}\quad{} \hat{l}_{i} = \Phi{}_{\textit{FADeR}}(\tau{}^{I'}_{i}) 
    \end{aligned}
    \label{eq:pred_fader}
\end{equation}

FADeR learns to predict patch-wise error considering the suspected defective rank among patches.
We employ ranking loss as ~\eqref{eq:loss_rank} to train patch-wise error prediction.
This allows the model to learn the relative numerical differences of the patch-wise reconstruction error.
For reference, training with conventional loss functions such as $\mathcal{L}_{2}$ focuses only on the absolute difference of each patch-wise error.
This conventional loss function is not recommended to train FADeR because includes possible side effect on training.
For example when some patches have similar reconstruction error as GT, FADeR possibly mislearn the degree of suspected defective and those attenuation rankings may be reversed.

\vspace{-0.3cm}
\begin{equation}
    \begin{aligned}
        \mathcal{L}_{\textit{mrank}}(\hat{l}, l) = \text{max}\left( 0, -\mathds{1}(l_{i}, l_{j}) \cdot{} (\hat{l}_{i}-\hat{l}_{j})+\xi \right) \\
        \text{s.t.}\quad{} \mathds{1}(l_{i}, l_{j}) = 
            \begin{cases}
                +1, \ \text{if}\ \  l_{i} > l_{j} \\
                -1, \ \text{otherwise}
            \end{cases}
    \end{aligned}
    \label{eq:loss_rank}
\end{equation}

\subsection{Soft mask for relative attenuation}
When the predicted values for patch-wise error show an equal level overall, it indicates an anomaly-free situation.
Conversely, when a significant disparity exists among the predicted patch-wise errors, we can decide there is a suspected defective representation to be masked.

Existing masking methods highly rely on binary mask $M_{\textit{B}}$. 
This binary masking method has a critical drawback.
In scenarios such as the above anomaly-free situation, unnecessary masked regions will be obligatorily generated for some patches that have relatively larger values than others.
Then, it will result the inaccurate normal reconstructions.

\begin{table*}[t]
    \centering
    \scriptsize
    \caption{Summary of the AUROC for the MVTec AD dataset~\cite{MVTec_Bergmann_CVPR19}. NNs are structured with simple well-known reconstruction backbones AE and U-Net~\cite{Unet_Olaf_MICCAI15}. Abbreviations of attention module, discriminator, and memory module are `Att', `Dis', and `Mem' respectively. FADeR$_{\textit{EAR}}$ represents FADeR combined with EAR.}
    \setlength\tabcolsep{4pt}
        \begin{tabular}{l || *{4}{P{1.15cm}} | *{4}{P{1.15cm}} | *{1}{P{1.17cm}}}
        \hline
            \multirow{2}{*}{Model} & MS-CAM & GANomaly & SCADN & MemAE & U-Net & DAAD & RIAD & EAR & FADeR$_{\textit{EAR}}$ \\
            & ~\cite{MSCAM_Li_Sensors22} & ~\cite{GANomaly_Akcay_ACCV18} & ~\cite{SCADN_Yan_AAAI21} & ~\cite{MemAE_Gong_ICCV19} & ~\cite{Unet_Olaf_MICCAI15} & ~\cite{BWMem_Hou_ICCV21} & ~\cite{RIAD_Vitzan_PR21} & ~\cite{EAR_Park_arXiv23} & (ours) \\
        \hline
            Backbone & \multicolumn{4}{c|}{AE} &\multicolumn{4}{c|}{U-Net} & U-Net \\
        \hline 
            Additional & \multirow{2}{*}{Att} & \multirow{2}{*}{Dis} & \multirow{2}{*}{Dis} & \multirow{2}{*}{Mem} & \multirow{2}{*}{-} & Dis & \multirow{2}{*}{-} & \multirow{2}{*}{-} & two-layer \\
            Module & & & & & & Mem & & & MLP \\
        \hline
        \hline
            Bottle & 0.940 & 0.892 & 0.957 & 0.930 & 0.863 & 0.976 & \textbf{0.999} & 0.997 & 0.998 \\
            Cable & 0.880 & 0.732 & 0.856 & 0.785 & 0.636 & 0.844 & 0.819 & 0.871 & \textbf{0.887} \\
            Capsule & 0.850 & 0.708 & 0.765 & 0.735 & 0.673 & 0.767 & 0.884 & 0.870 & \textbf{0.947} \\
            Carpet & 0.910 & 0.842 & 0.504 & 0.386 & 0.774 & 0.866 & 0.842 & 0.899 & \textbf{0.971} \\
            Grid & 0.940 & 0.743 & 0.983 & 0.805 & 0.857 & 0.957 & \textbf{0.996} & 0.959 & 0.983 \\
            Hazelnut & 0.950 & 0.794 & 0.833 & 0.769 & 0.996 & 0.921 & 0.833 & \textbf{0.997} & 0.988 \\
            Leather & 0.950 & 0.792 & 0.659 & 0.423 & 0.870 & 0.862 & \textbf{1.000} & \textbf{1.000} & \textbf{1.000} \\
            Metal nut & 0.690 & 0.745 & 0.624 & 0.654 & 0.676 & 0.758 & 0.885 & \textbf{0.876} & \textbf{0.876} \\
            Pill & 0.890 & 0.757 & 0.814 & 0.717 & 0.781 & 0.900 & 0.838 & 0.922 & \textbf{0.976} \\
            Screw & \textbf{1.000} & 0.699 & 0.831 & 0.257 & \textbf{1.000} & 0.987 & 0.845 & 0.886 & 0.918 \\
            Tile & 0.800 & 0.785 & 0.792 & 0.718 & 0.964 & 0.882 & 0.987 & 0.965 & \textbf{1.000} \\
            Toothbrush & \textbf{1.000} & 0.700 & 0.891 & 0.967 & 0.811 & 0.992 & \textbf{1.000} & \textbf{1.000} & \textbf{1.000} \\
            Transistor & 0.880 & 0.746 & 0.863 & 0.791 & 0.674 & 0.876 & 0.909 & \textbf{0.947} & 0.933 \\
            Wood & 0.940 & 0.653 & 0.968 & 0.954 & 0.958 & 0.982 & 0.930 & 0.985 & \textbf{0.996} \\
            Zipper & 0.910 & 0.834 & 0.846 & 0.710 & 0.750 & 0.859 & 0.981 & 0.955 & \textbf{0.987} \\
        \hline
        \hline
            Average & 0.902 & 0.761 & 0.812 & 0.707 & 0.819 & 0.895 & 0.917 & 0.942 & \textbf{0.964} \\
        \hline
        \end{tabular}
    \label{table:perform_mvtec}
    \vspace{-0.3cm}
\end{table*}

To resolve the issue of binary masking, we propose a soft masking strategy.
We first normalize predicted patch-wise error values $\hat{l}$ into range $(0, 1)$.
Then, flip the value of $\hat{l}$ into $(1, 0)$.
We call it a soft mask and mark it with $M_{\textit{S}}$.
Then, we apply $M_{\textit{S}}$ to each skip connection as~\eqref{eq:attenuation}.

\vspace{-0.2cm}
\begin{equation}
    \begin{aligned}
        \Phi{}_{\textit{Dec}}^{d}\left( \text{concat}(\  \Phi{}_{\textit{Dec}}^{d+1}, \Phi{}_{\textit{Enc}}^{d} \odot{} \Theta(M_{\textit{S}}(\hat{l}))\  ) \right) \\
        \text{w.r.t.}\quad{} 1\leq{}d\leq{}D-1
    \end{aligned}
    \label{eq:attenuation}
\end{equation}

The depth level of the U-Net is represented with $d$ with the maximum depth $D$. 
The deepest block of this U-Net does not have a skip connection so FADeR is applied between the $1$ to $D-1$.
To cope with the resolution of $\Phi{}_{\textit{Enc}}^{d}$ that varies depending on the depth of the U-Net, we use the scaling function $\Theta$.
In contrast to the binary masking method, this soft masking method prevents unintended cut-out situations of normal feature representations. 
In addition, it effectively prevents an IS issue by greatly attenuating defective representation. 

\section{Experiments}

\begin{figure}[t]
    \resizebox{\columnwidth}{!}{%
        \begin{tabular}{c@{\hspace{0.1cm}}c@{\hspace{0.1cm}}c@{\hspace{0.1cm}}c}
            Input & DINO-ViT~\cite{DINO_Caron_ICCV21} & Plain-ViT~\cite{ViT_Alexey_ICLR21} & WinCLIP~\cite{WinCLIP_Jeong_CVPR23} \\
            \includegraphics*[width=0.29\columnwidth,trim={0.0cm 0.0cm 0.0cm 0.0cm},clip]{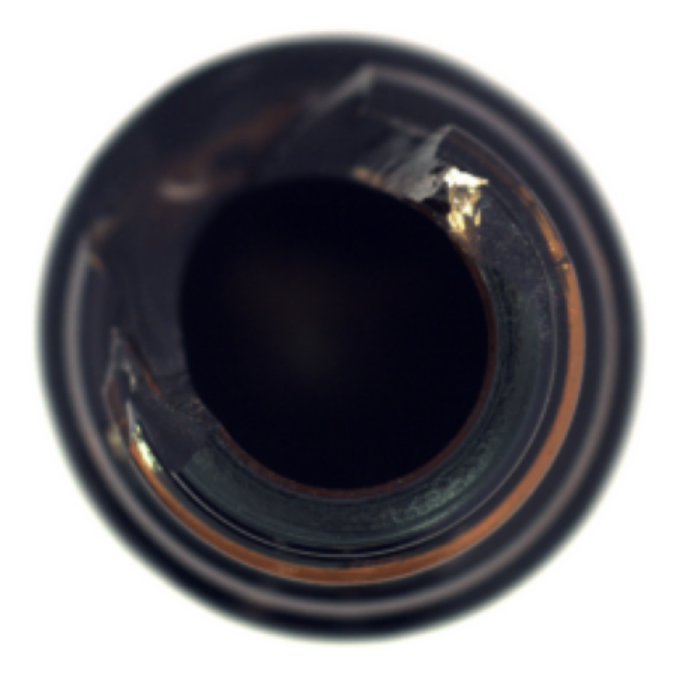} &
            \includegraphics*[width=0.29\columnwidth,trim={0.0cm 0.0cm 0.0cm 0.0cm},clip]{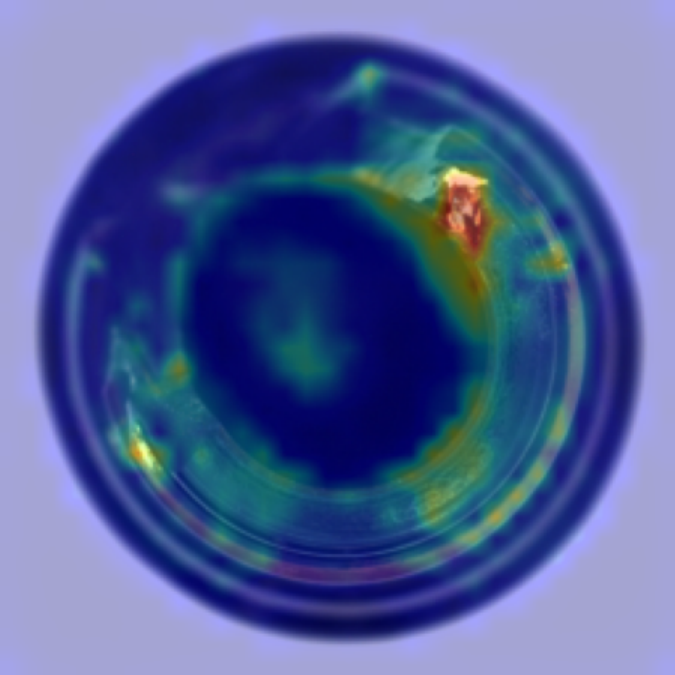} &
            \includegraphics*[width=0.29\columnwidth,trim={0.0cm 0.0cm 0.0cm 0.0cm},clip]{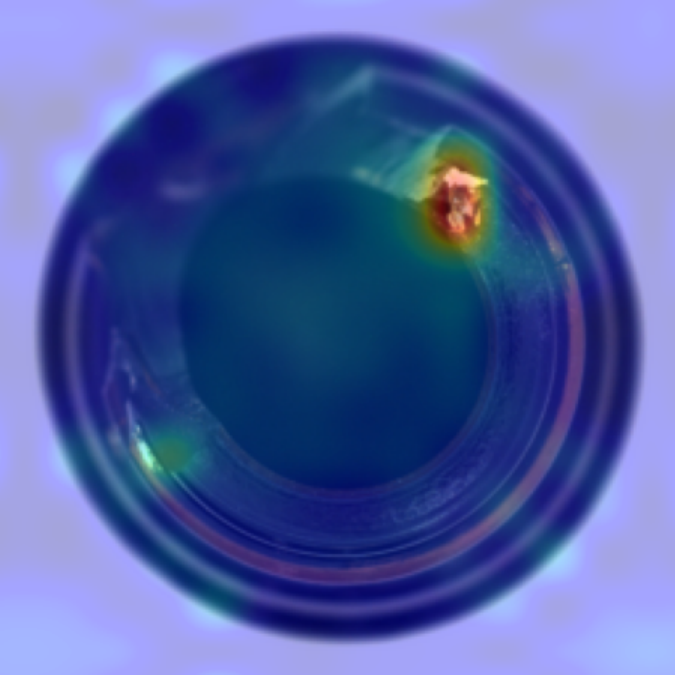} &
            \includegraphics*[width=0.29\columnwidth,trim={0.0cm 0.0cm 0.0cm 0.0cm},clip]{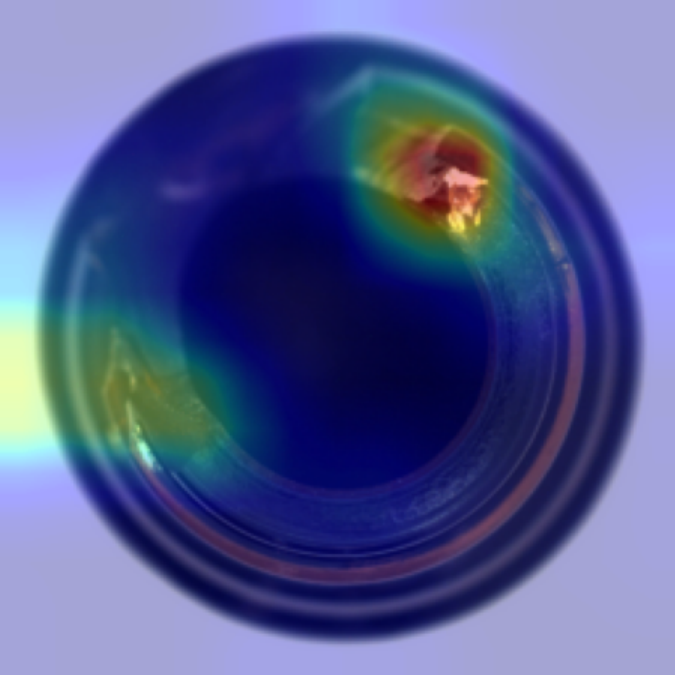} \\

            \includegraphics*[width=0.29\columnwidth,trim={0.0cm 0.0cm 0.0cm 0.0cm},clip]{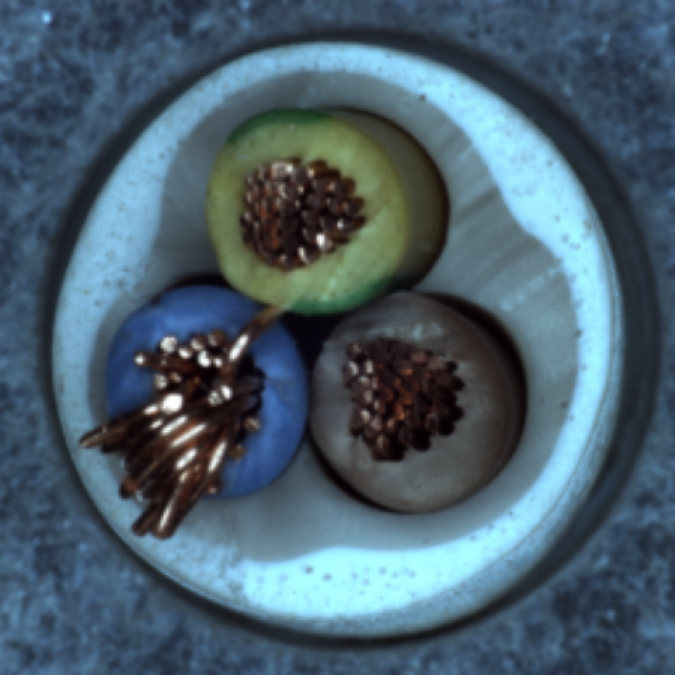} &
            \includegraphics*[width=0.29\columnwidth,trim={0.0cm 0.0cm 0.0cm 0.0cm},clip]{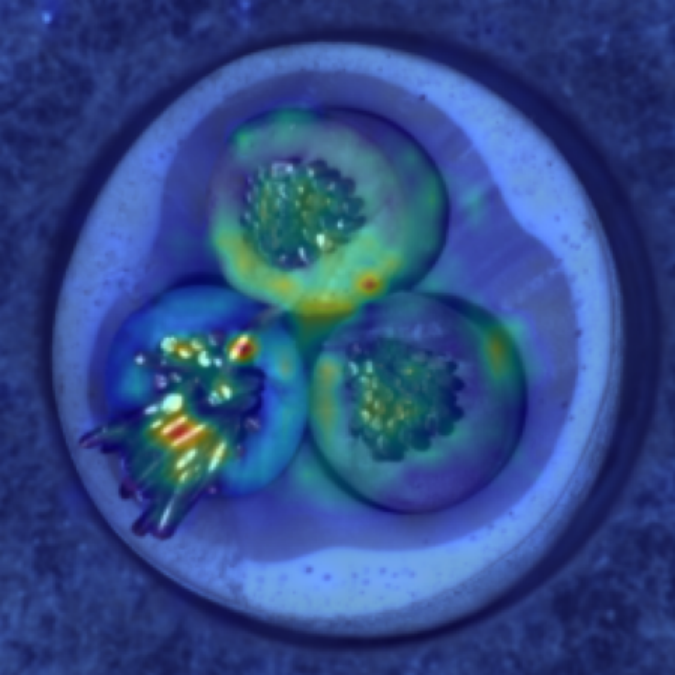} &
            \includegraphics*[width=0.29\columnwidth,trim={0.0cm 0.0cm 0.0cm 0.0cm},clip]{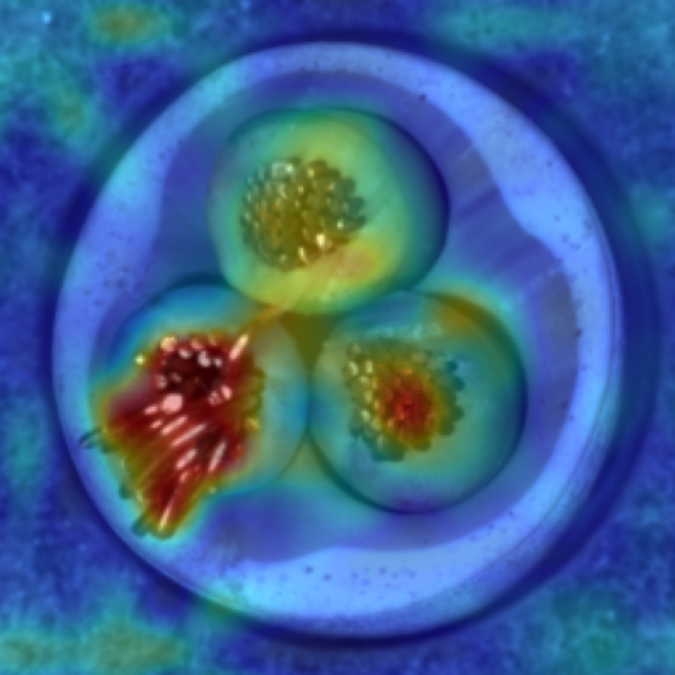} &
            \includegraphics*[width=0.29\columnwidth,trim={0.0cm 0.0cm 0.0cm 0.0cm},clip]{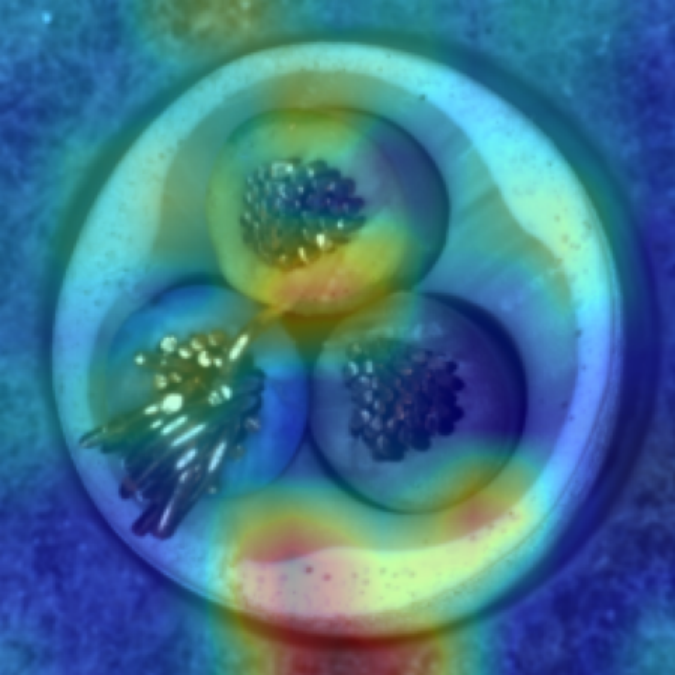} \\

            \includegraphics*[width=0.29\columnwidth,trim={0.0cm 0.0cm 0.0cm 0.0cm},clip]{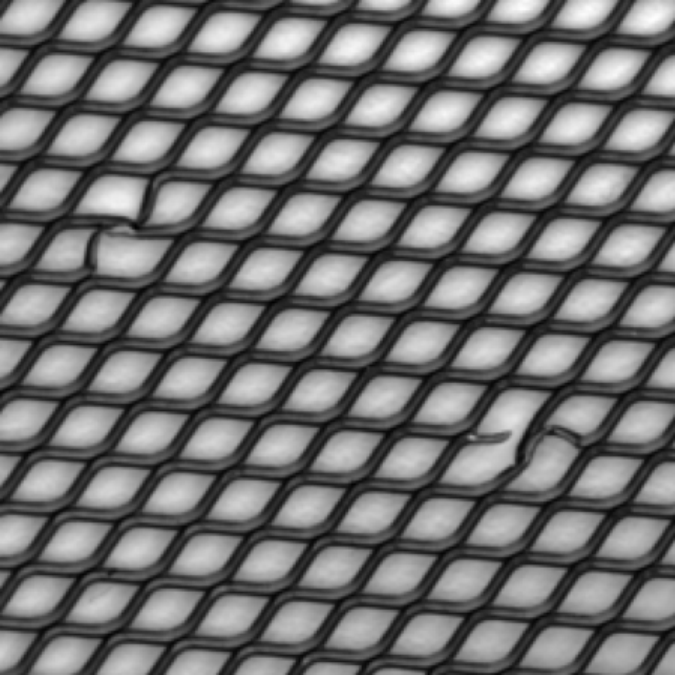} &
            \includegraphics*[width=0.29\columnwidth,trim={0.0cm 0.0cm 0.0cm 0.0cm},clip]{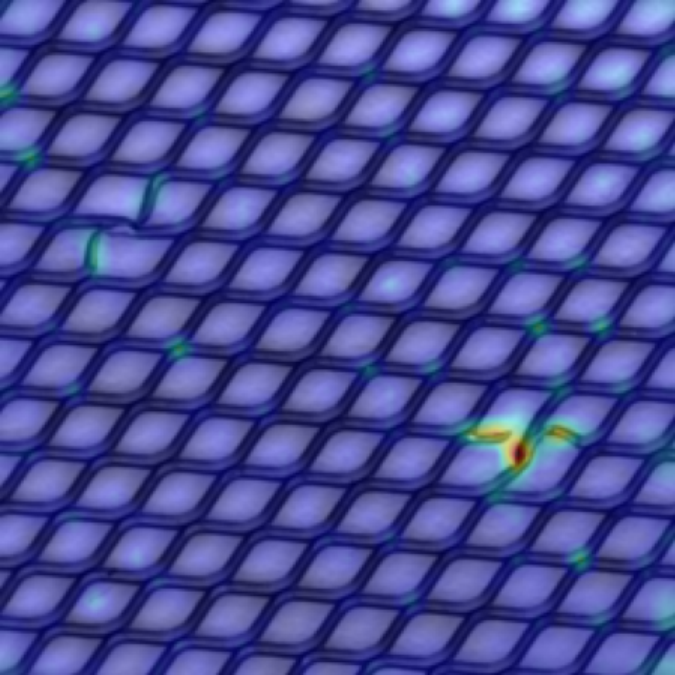} &
            \includegraphics*[width=0.29\columnwidth,trim={0.0cm 0.0cm 0.0cm 0.0cm},clip]{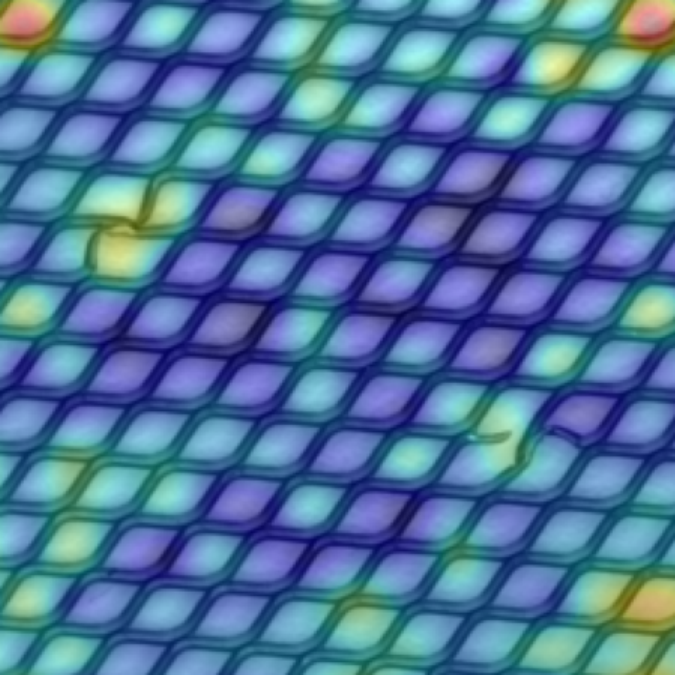} &
            \includegraphics*[width=0.29\columnwidth,trim={0.0cm 0.0cm 0.0cm 0.0cm},clip]{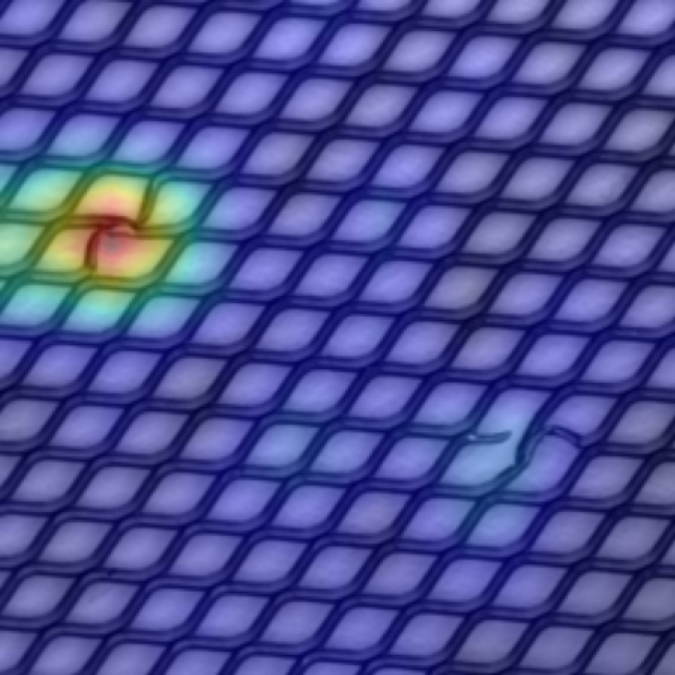} \\

            \includegraphics*[width=0.29\columnwidth,trim={0.0cm 0.0cm 0.0cm 0.0cm},clip]{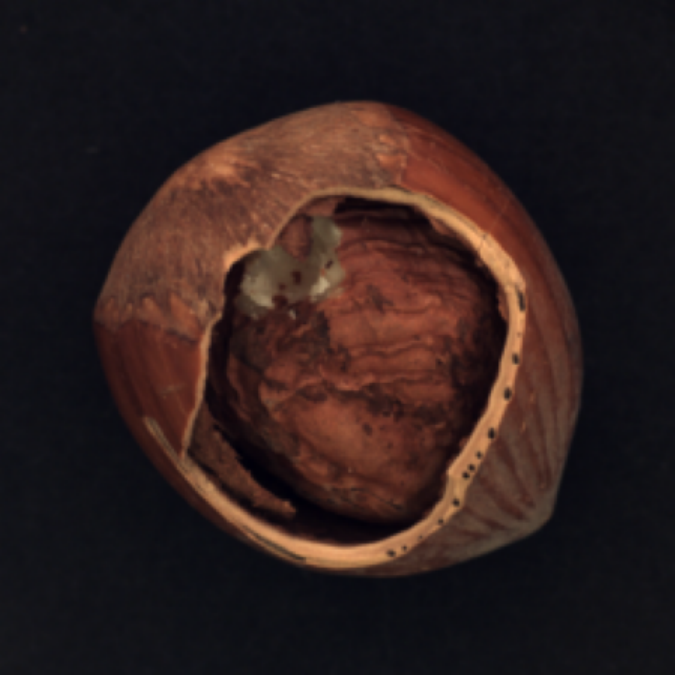} &
            \includegraphics*[width=0.29\columnwidth,trim={0.0cm 0.0cm 0.0cm 0.0cm},clip]{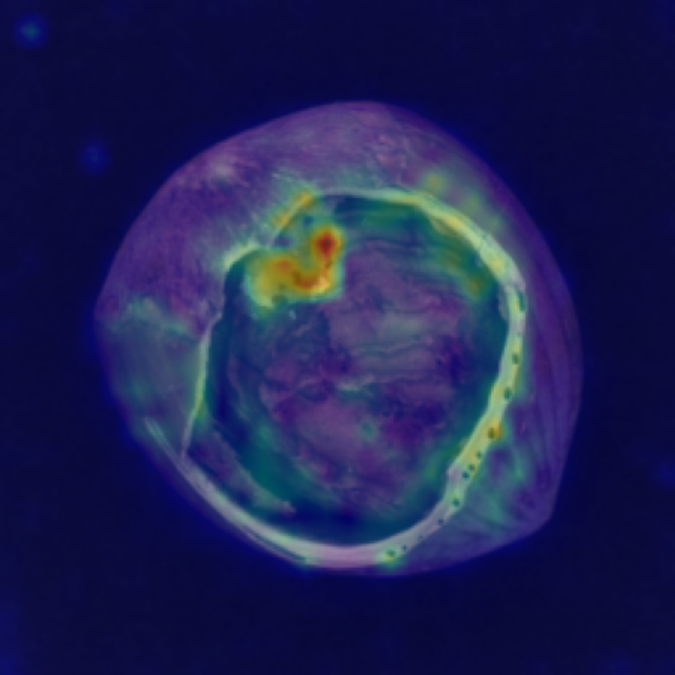} &
            \includegraphics*[width=0.29\columnwidth,trim={0.0cm 0.0cm 0.0cm 0.0cm},clip]{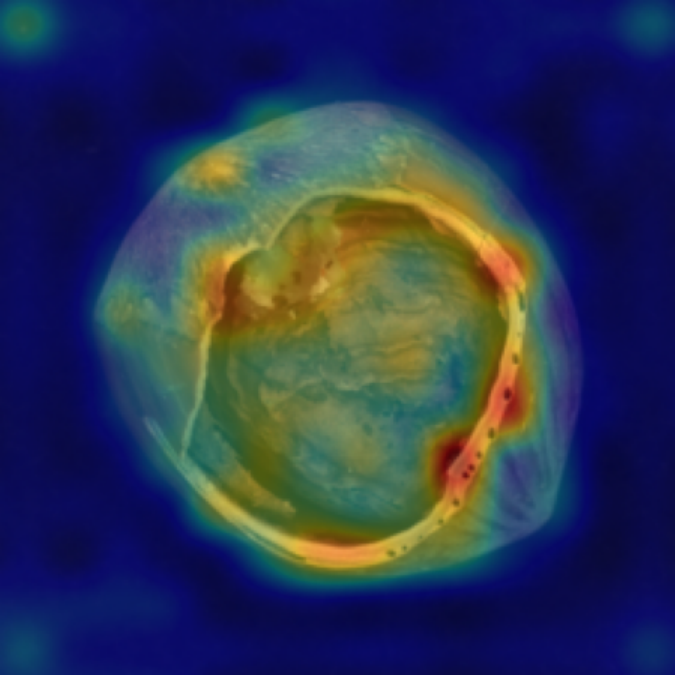} &
            \includegraphics*[width=0.29\columnwidth,trim={0.0cm 0.0cm 0.0cm 0.0cm},clip]{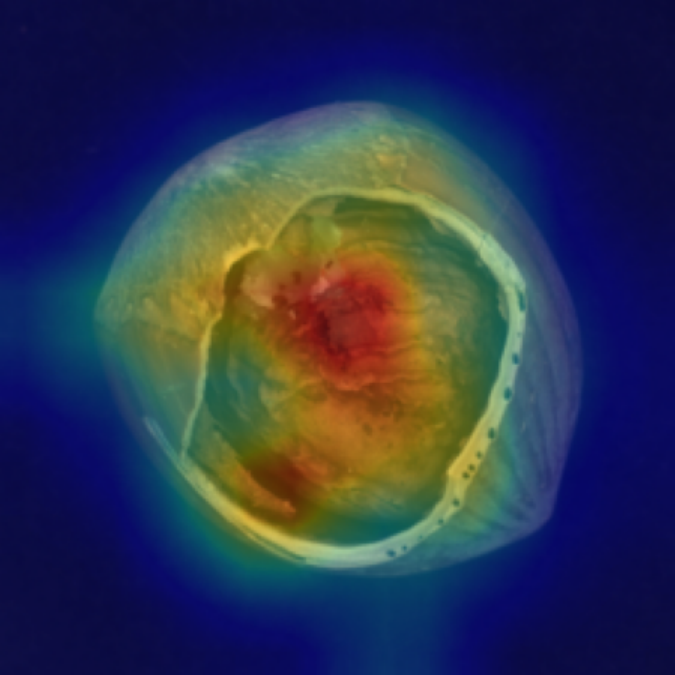} \\

            \includegraphics*[width=0.29\columnwidth,trim={0.0cm 0.0cm 0.0cm 0.0cm},clip]{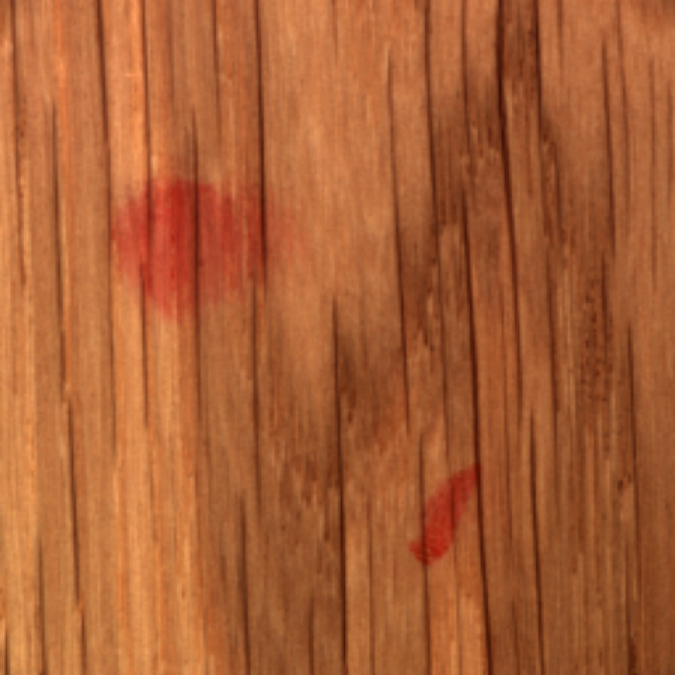} &
            \includegraphics*[width=0.29\columnwidth,trim={0.0cm 0.0cm 0.0cm 0.0cm},clip]{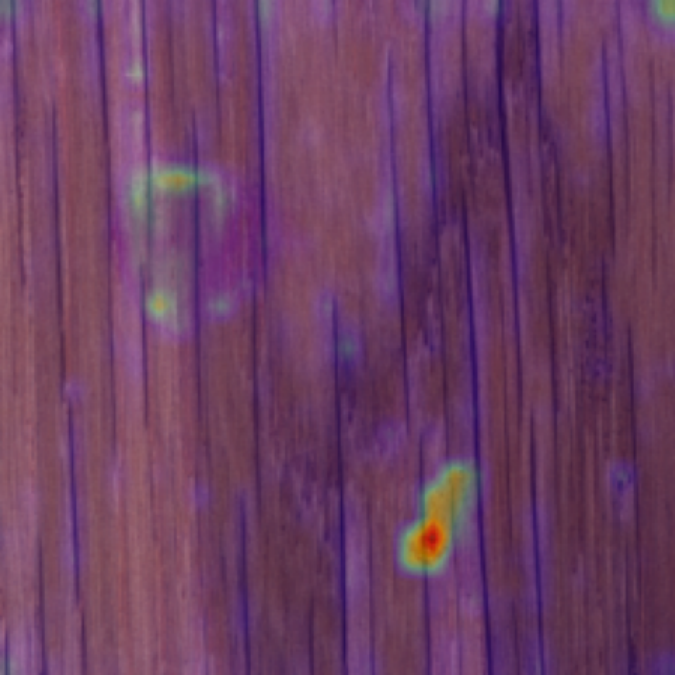} &
            \includegraphics*[width=0.29\columnwidth,trim={0.0cm 0.0cm 0.0cm 0.0cm},clip]{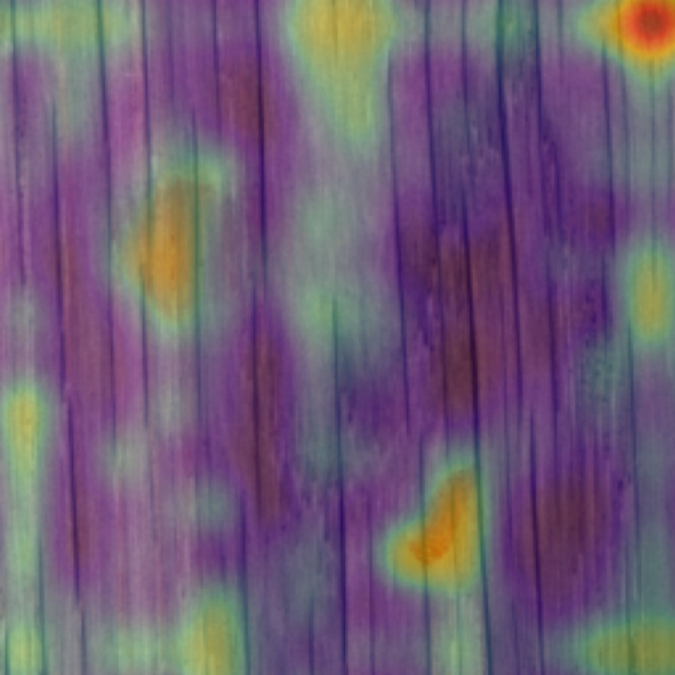} &
            \includegraphics*[width=0.29\columnwidth,trim={0.0cm 0.0cm 0.0cm 0.0cm},clip]{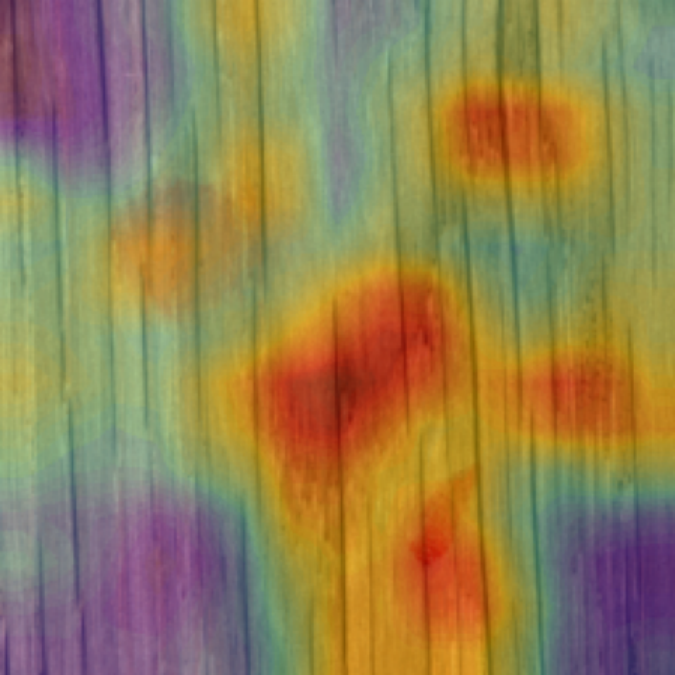} \\
        \end{tabular}
    }
    \vspace{-0.2cm}
    \caption{Results of attention map generation on various pre-trained attention. Note that, the pre-trained attention pays attention to visual saliency regions that are distinguished from surrounding information. By exploiting this property, we can generate a mask of suspected defective regions but the results of the binary mask will be different depending on the attention mechanism.}
    \label{fig:attention_missing}
\end{figure}

\subsection{Experimental setup}
To evaluate our method, we use two public industrial visual inspection datasets MVTec AD~\cite{MVTec_Bergmann_CVPR19} and VisA~\cite{VisA_Zou_ECCV22}.
MVTec AD~\cite{MVTec_Bergmann_CVPR19} provides a total of 15 subtasks with 10 objects and 5 textures.
Another dataset, VisA~\cite{VisA_Zou_ECCV22}, provides 12 different objects including cases where multiple objects exist in a single image.
All the training processes are performed with only normal samples.
The UAD performance is evaluated by both normal and defective samples.

\textbf{Implementation details.}
On the line of succession EAR~\cite{EAR_Park_arXiv23}, our goal is to achieve better UAD performance while avoiding increasing the scale of NNs.
We also avoid scratch learning an NN for a given task.
For this, we attach FADeR on the pre-trained reconstruction-by-inpainting model EAR~\cite{EAR_Park_arXiv23}.
The pre-trained U-Net~\cite{Unet_Olaf_MICCAI15} that we used includes five convolutional blocks ($D=5$) for each encoder and decoder.
The layer of $d$-th depth in the encoder is concatenated to the same level decoder via skip connection except for the deepest layer.

For each encoder block, we repeat `convolution $\rightarrow$ batch normalization $\rightarrow$ leaky ReLU activation' three times.
In the third layer of each encoder block, the stride value is set to 2 for downscaling.
The decoder is constructed by repeating `upsampling $\rightarrow$ convolution $\rightarrow$ batch normalization $\rightarrow$ leaky ReLU activation' three times.
Note that, the upscaling in the decoding process is performed with the nearest interpolation by the scaling factor 2.

\textbf{Training conditions.}
In EAR~\cite{EAR_Park_arXiv23}, U-Net~\cite{Unet_Olaf_MICCAI15} was pre-trained on the MVTec AD dataset~\cite{MVTec_Bergmann_CVPR19}.
EAR also provides an optimal mosaic scale $m^{*}$ for visual obfuscation-based hint providing.
We exploit the $m^{*}$ as the patch size for the tokenization process in FADeR as aforementioned.
FADeR is trained for 100 epochs with only normal samples.
The training process includes a hyperparameter tuning only for conditions related to the learning rate $\eta$ as follows: 
1) 1e-3, 1e-4, and 1e-5 for $\eta$ and 
2) fixed $\eta$, learning rate warm-up~\cite{Warm_Goyal_arXiv17}, and SGDR~\cite{SGDR_Ilya_ICLR17} for learning rate scheduling.
While training FADeR we also conduct fine-tuning of the decoder.
We set the learning rate for fine-tuning to the $0.1 \eta$.
This fine-tuning ensures an accurate reconstruction of normal patterns, preserving contained generalization ability.

\textbf{Evaluation metric.}
To evaluate the performance of UAD experiments, the area under the receiver operating characteristic curve (AUROC)~\cite{AUROC_Tom_PRL06} is used.
The AUROC is measured based on the anomaly scores for each normal and defective sample within the test set at both the image level and the pixel level.
The anomaly scoring function is constructed with multi-scale gradient magnitude similarity (MSGMS)~\cite{RIAD_Vitzan_PR21} that allows the detection of various sizes of defects.
The anomaly score of the UAD model for the unseen anomalous patterns will be close to 1 or relatively larger compared to the seen normal cases.

\subsection{Visual analysis of soft mask}
The visual comparison between the binary mask and soft mask is shown in Figure~\ref{fig:vis_results}.
The first and the second columns show the input images and the binary masks over them respectively.
Those binary masks are generated from the ImageNet pre-trained DINO-VIT~\cite{DINO_Caron_ICCV21} attention map.
The last third column shows the soft mask of FADeR. 
We observed incomplete mask cases for binary attention mask around the insulation tearing on the right side of the cable and the broken leg of the transistor. 
This incomplete mask problem of defective regions is resolved by exploiting the soft mask of FADeR as shown in the third column. 
In the soft mask $M_{\textit{S}}$, the defective representation will be greatly attenuated during transmission via skip connection of U-Net.
This prevents accurate reconstruction of the unseen defective patterns and an IS issue of U-Net.

In case of the capsule in Figure~\ref{fig:vis_results}, the defective region is fully covered by a binary attention mask. 
When the mask is larger than the defective region, it hinders the reconstruction of normal patterns.
Moreover, the EAR method may also give an unnecessary hint of defective representations.
On the other hand, the soft mask of FADeR recognizes defective regions with an appropriate mask size than the binary mask as shown in the figure.
This helps cut out the defective representation and unnecessary hints on defective regions.

\subsection{Performance of industrial inspection}
In this section, we quantitatively evaluate our method by comparing existing SOTA methods. 
We value the importance of edge computing capabilities in the industry.
Thus, our goal is to maximize the UAD performance without significantly increasing the scale of NNs. 
To guarantee the quality of the product, the highest priority is to determine whether given product is acceptable or not at the manufacturing site.
Therefore, we compare the image-level AUROC.
The models to compare the performance are based on U-Net~\cite{Unet_Olaf_MICCAI15} or similar scales of NNs.
The measured performance with MVTec AD dataset~\cite{MVTec_Bergmann_CVPR19} is summarized in Table~\ref{table:perform_mvtec}.

FADeR is a two-layer MLP attached to pre-trained EAR~\cite{EAR_Park_arXiv23}. 
Comparing the performance with the plain EAR case, adding FADeR achieves notable performance improvements in the capsule, carpet, metal nut, and pill cases. 
In the hazelnut and transistor cases, the performance slightly decreases compared to the plain EAR, but it is still comparable to or better than the other competitive methods. 
Overall, FADeR achieves the best performance in 10 out of the 15 subtasks on the MVTec AD dataset~\cite{MVTec_Bergmann_CVPR19}, with an average AUROC of 0.964.

\begin{table}[t]
    \centering
    \scriptsize
    \caption{Performance of UAD and defect localization in terms of image/pixel level AUROC for three pre-trained attentions. Each left and right side of ` $\rightarrow$ ' represents before and after applying FADeR on EAR~\cite{EAR_Park_arXiv23}. Results are shown for pre-trained attention of DINO-ViT~\cite{DINO_Caron_ICCV21}, Plain-ViT~\cite{ViT_Alexey_ICLR21}, and WinCLIP~\cite{WinCLIP_Jeong_CVPR23}.}
    \setlength\tabcolsep{1pt}
    \resizebox{\columnwidth}{!}{%
    \renewcommand{\arraystretch}{1.5}
    \begin{tabular}{l || *{1}{P{3.11cm}} | *{1}{P{3.11cm}} | *{1}{P{3.11cm}}}
        \hline
            Model & FADeR$_{\textit{DINO-EAR}}$ & FADeR$_{\textit{ViT-EAR}}$ & FADeR$_{\textit{WinCLIP-EAR}}$ \\
        \hline
        \hline
            Bottle & 0.997 / 0.914 $\rightarrow$ \textbf{0.998 / 0.951} & 0.983 / 0.930 $\rightarrow$ \textbf{0.993 / 0.950} & \textbf{0.992} / 0.939 $\rightarrow$ \textbf{0.992 / 0.936} \\
            Cable & 0.870 / 0.775 $\rightarrow$ \textbf{0.887 / 0.856} & 0.817 / 0.798 $\rightarrow$ \textbf{0.825 / 0.872} & \textbf{0.838} / 0.550 $\rightarrow$ 0.794 / \textbf{0.612} \\
            Capsule & 0.870 / 0.944 $\rightarrow$ \textbf{0.947 / 0.980} & 0.863 / 0.965 $\rightarrow$ \textbf{0.933 / 0.977} & 0.867 / 0.935 $\rightarrow$ \textbf{0.930 / 0.968} \\
            Carpet & 0.899 / 0.974 $\rightarrow$ \textbf{0.971 / 0.992} & 0.701 / 0.945 $\rightarrow$ \textbf{0.756 / 0.981} & 0.679 / 0.909 $\rightarrow$ \textbf{0.746 / 0.955} \\
            Grid & 0.959 / 0.974 $\rightarrow$ \textbf{0.983 / 0.986} & 0.765 / \textbf{0.919} $\rightarrow$ \textbf{0.805} / 0.901 & \textbf{0.866 / 0.940} $\rightarrow$ 0.861 / 0.939 \\
            Hazelnut & 0.997 / 0.957 $\rightarrow$ \textbf{0.988 / 0.976} & 0.954 / 0.952 $\rightarrow$ \textbf{0.973 / 0.977} & 0.946 / 0.947 $\rightarrow$ \textbf{0.959 / 0.976} \\
            Leather & \textbf{1.000} / 0.992 $\rightarrow$ \textbf{1.000 / 0.996} & \textbf{1.000} / 0.981 $\rightarrow$ \textbf{1.000 / 0.993} & \textbf{1.000} / 0.971 $\rightarrow$ \textbf{1.000 / 0.990} \\
            Metal nut & \textbf{0.876} / 0.793 $\rightarrow$ \textbf{0.876 / 0.828} & 0.880 / 0.767 $\rightarrow$ \textbf{0.903 / 0.804} & 0.775 / 0.780 $\rightarrow$ \textbf{0.821 / 0.779} \\
            Pill & 0.922 / 0.875 $\rightarrow$ \textbf{0.976 / 0.945} & 0.884 / 0.914 $\rightarrow$ \textbf{0.957 / 0.967} & 0.879 / 0.898 $\rightarrow$ \textbf{0.956 / 0.951} \\
            Screw & 0.886 / 0.975 $\rightarrow$ \textbf{0.918 / 0.991} & 0.883 / 0.977 $\rightarrow$ \textbf{0.905 / 0.993} & 0.862 / 0.953 $\rightarrow$ \textbf{0.897 / 0.968} \\
            Tile & 0.962 / 0.857 $\rightarrow$ \textbf{1.000 / 0.950} & 0.961 / 0.853 $\rightarrow$ \textbf{1.000 / 0.940} & 0.960 / 0.816 $\rightarrow$ \textbf{0.986 / 0.921} \\
            Toothbrush & \textbf{1.000} / 0.953 $\rightarrow$ \textbf{1.000 / 0.987} & \textbf{1.000} / 0.953 $\rightarrow$ 0.994 / \textbf{0.986} & \textbf{1.000} / 0.959 $\rightarrow$ \textbf{1.000 / 0.986} \\
            Transistor & \textbf{0.947} / 0.745 $\rightarrow$ 0.933 / \textbf{0.825} & 0.891 / 0.679 $\rightarrow$ \textbf{0.908 / 0.740} & 0.788 / \textbf{0.699} $\rightarrow$ \textbf{0.835} / 0.642 \\
            Wood & 0.985 / \textbf{0.875} $\rightarrow$ 0.996 / \textbf{0.818} & 0.931 / \textbf{0.818} $\rightarrow$ \textbf{0.964} / 0.730 & \textbf{0.997} / 0.691 $\rightarrow$ \textbf{0.997 / 0.751} \\
            Zipper & 0.955 / 0.930 $\rightarrow$ \textbf{0.987 / 0.988} & 0.932 / 0.909 $\rightarrow$ \textbf{0.938 / 0.951} & 0.924 / 0.895 $\rightarrow$ \textbf{0.975 / 0.967} \\
        \hline
        \hline
            Average & 0.942 / 0.902 $\rightarrow$ \textbf{0.964 / 0.938} & 0.896 / 0.891 $\rightarrow$ \textbf{0.924 / 0.918} & 0.892 / 0.859 $\rightarrow$ \textbf{0.917 / 0.889} \\
        \hline
    \end{tabular}
    \renewcommand{\arraystretch}{1}
    }
    \label{table:perform_attns}
\end{table}

\subsection{Generalizing FADeR on other attentions}
The characteristic of attention depends on the NN structure and the training dataset~\cite{AGDM_Lee_Bigcomp24}.
From the visual comparison shown in Figure~\ref{fig:attention_missing}, we can see that the region of interest and its intensity vary depending on the pre-trained attention model. 

Our purpose is to overcome the possible incomplete mask issue in binary attention masks. 
To generalize the effectiveness of FADeR in various pre-trained attention other than DINO-ViT~\cite{DINO_Caron_ICCV21}, we also apply FADeR to the original plain ViT (Plain-ViT)~\cite{ViT_Alexey_ICLR21} and WinCLIP~\cite{WinCLIP_Jeong_CVPR23}. 
WinCLIP~\cite{WinCLIP_Jeong_CVPR23} features a zero-shot defect detection by highlighting suspected anomalous regions by leveraging text prompts of defective descriptions.

In this experiment, we additionally measure pixel-level AUROC to confirm that the soft mask resolves the incomplete mask issue of defective regions.
An increase in pixel-level AUROC indicates that precise defective recognition at pixel-level. 
The measured image-level and pixel-level AUROC with MVTec AD dataset~\cite{MVTec_Bergmann_CVPR19} is summarized in Table~\ref{table:perform_attns}.

DINO-ViT~\cite{DINO_Caron_ICCV21} is known to give strong attention to patterns that are notably different from their surroundings~\cite{DINO_Caron_ICCV21}.
When it used for AD purposes, this feature is advantageous to make strong attention to suspected defective regions that are different from surrounding normal patterns.
By leveraging this, we generate attention masks with DINO-ViT~\cite{DINO_Caron_ICCV21} to cover defective regions in a zero-shot manner.
In contrast, the other two attention models show different attention map compared to DINO-ViT~\cite{DINO_Caron_ICCV21}.
Thus, the incomplete masking cases will also vary when changing the pre-trained attention model for single deterministic mask generation.
In addition, it affects the UAD performance.
As a result, We confirm that EAR~\cite{EAR_Park_arXiv23} based on DINO-ViT~\cite{DINO_Caron_ICCV21} achieves the best AUROC.
On the other hand, EAR with other attention cases shows degraded UAD performances because their low-contrast attention makes a large mask.

The results of applying FADeR are shown to the right side of the '$\rightarrow$' in Table~\ref{table:perform_attns}.
The image-level or pixel-level AUROC was increased in 44 out of 45 independent subtasks shown in Table~\ref{table:perform_attns}, with average increases 2.90\% and 3.6\%, respectively. 
The increase in pixel-level AUROC suggests that the soft mask effectively cuts out defective representation in skip connection.

\begin{table}[t]
    \centering
    \footnotesize
    \caption{Results of the ablation study. All variants are set to apply FADeR on EAR~\cite{EAR_Park_arXiv23}. We deal with three conditions 1) binary masking or soft masking, 2) interpolation method for mask scaling, and 3) NN structure for FADeR. }
    \setlength\tabcolsep{1.5pt}
    \resizebox{\columnwidth}{!}{%
    \begin{tabular}{l || *{1}{P{1.4cm}} | *{1}{P{1.4cm}} | *{1}{P{1.4cm}} | *{1}{P{1.4cm}} | *{1}{P{1.4cm}}}
        \hline
            Model & EAR~\cite{EAR_Park_arXiv23} & \multicolumn{4}{c}{FADeR$_{\textit{EAR}}$} \\
        \hline
        \hline
            Structure & U-Net & $+$ViT-B/16 & \multicolumn{3}{c}{$+$two-layer MLP}\\
        \hline
            Masking target & Image & \multicolumn{4}{c}{Feature map} \\
        \hline
            Masking & Hard & Soft & Hard & \multicolumn{2}{c}{Soft} \\
        \hline
            Scaling & - & \multicolumn{2}{c|}{Nearest} & Bilinear & Nearest \\
        \hline
        \hline
            Bottle & 0.997 & \textbf{0.998} & 0.997 & \textbf{0.998} & \textbf{0.998} \\
            Cable & 0.871 & 0.878 & \textbf{0.900} & 0.893 & 0.887 \\
            Capsule & 0.870 & 0.941 & 0.939 & 0.946 & \textbf{0.947} \\
            Carpet & 0.899 & \textbf{0.985} & 0.664 & 0.973 & 0.971 \\
            Grid & 0.959 & 0.982 & 0.794 & \textbf{0.983} & \textbf{0.983} \\
            Hazelnut & 0.997 & 0.983 & 0.970 & 0.980 & \textbf{0.988} \\
            Leather & \textbf{1.000} & \textbf{1.000} & \textbf{1.000} & \textbf{1.000} & \textbf{1.000} \\
            Metal nut & 0.876 & 0.875 & 0.817 & \textbf{0.879} & 0.876 \\
            Pill & 0.922 & 0.972 & 0.636 & 0.974 & \textbf{0.976} \\
            Screw & 0.886 & \textbf{0.923} & 0.605 & 0.914 & 0.918 \\
            Tile & 0.965 & \textbf{1.000} & 0.799 & \textbf{1.000} & \textbf{1.000} \\
            Toothbrush & \textbf{1.000} & \textbf{1.000} & 0.989 & \textbf{1.000} & \textbf{1.000} \\
            Transistor & \textbf{0.947} & 0.905 & 0.894 & 0.899 & 0.933 \\
            Wood & 0.985 & 0.980 & 0.896 & 0.992 & \textbf{0.996} \\
            Zipper & 0.955 & \textbf{0.990} & 0.859 & 0.986 & 0.987 \\
        \hline
        \hline
            Average & 0.942 & 0.961 & 0.850 & 0.961 & \textbf{0.964} \\
        \hline
    \end{tabular}
    }
    \label{table:ablation_fader}
    \vspace{-0.2cm}
\end{table}

\subsection{Ablation study}
We can conduct an ablation study for FADeR on
1) a NN structure, 
2) a masking method (binary or soft masking), 
3) a mask scaling method to apply to a feature map, and
4) an NN to predict patch-wise error for mask generation.
FADeR employs a two-layer MLP to minimize the increase in computational complexity of the NN.
It predicts the patch-wise reconstruction error independently of surrounding patch information. 
To confirm the positive effect of patch-wise error prediction when reflecting the relationships between patches, we explore the ViT alter to MLP.
To demonstrate the effectiveness of the proposed method, we experiment with varying the remaining three conditions.

The advantage of using ViT lies in its capability to perform patch-wise error prediction based on understanding inter-patch dependencies through multi-head self-attention.
However, there is a marginal performance drop compared to ours in the ViT case as shown in the second column of Table~\ref{table:ablation_fader}. 
We confirm that the use of a two-layer MLP is the most effective considering the unnecessity of understanding surrounding information for predicting each patch-wise reconstruction error, along with ViT's computational complexity and the AUROC decrease.

As shown in the third column, adopting binary masking compared to soft masking makes degrades the UAD performance.
Because it performs unnecessary binary feature masking on normal patches.
When the soft mask is utilized with bilinear interpolation, some defective representations will leak via a smoothed patch border.
Due to this, the case of the bilinear interpolation shows a marginal performance drop compared to the best-performing case, as shown in the fourth column.

\begin{table}[t]
    \centering
    \scriptsize
    \caption{Performance of UAD and defect localization in terms of image/pixel level AUROC for other single masking methods. We replace a masking method based on pre-trained DINO-ViT~\cite{DINO_Caron_ICCV21} with DSR~\cite{DSR_ZavrtanikK_ECCV22}, MSFlow~\cite{MSFlow_Zhou_TNNLS24}, or AMI-Net~\cite{AMINet_Luo_TASC24}. These are marked as FADeR$_{\textit{DSR-EAR}}$, FADeR$_{\textit{MSFlow-EAR}}$, and FADeR$_{\textit{AMI-EAR}}$.}
    \setlength\tabcolsep{1pt}
    \resizebox{\columnwidth}{!}{%
    \renewcommand{\arraystretch}{1.5}
    \begin{tabular}{l || *{1}{P{3.11cm}} | *{1}{P{3.11cm}} | *{1}{P{3.11cm}}}
        \hline
            Model & FADeR$_{\textit{DSR-EAR}}$ & FADeR$_{\textit{MSFlow-EAR}}$ & FADeR$_{\textit{AMI-EAR}}$ \\
        \hline
        \hline
            Bottle  & \textbf{0.991 / 0.953} $\rightarrow$ 0.984 / 0.938 & 0.971 / 0.931 $\rightarrow$ \textbf{0.973 / 0.951} & 0.886 / 0.910 $\rightarrow$ \textbf{0.987 / 0.978} \\
            Cable   & \textbf{0.891} / 0.908 $\rightarrow$ 0.834 / \textbf{0.947} & 0.920 / 0.898 $\rightarrow$ \textbf{0.939 / 0.943} & \textbf{0.879} / 0.794 $\rightarrow$ 0.856 / \textbf{0.928} \\
            Capsule & 0.834 / 0.943 $\rightarrow$ \textbf{0.927 / 0.976} & 0.861 / 0.957 $\rightarrow$ \textbf{0.938 / 0.976} & \textbf{0.911} / 0.773 $\rightarrow$ 0.902 / \textbf{0.970} \\
            Carpet  & 0.806 / 0.982 $\rightarrow$ \textbf{0.935 / 0.993} & 0.983 / 0.970 $\rightarrow$ \textbf{0.984 / 0.982} & \textbf{0.983 / 0.932} $\rightarrow$ 0.785 / 0.903 \\
            Grid    & 0.930 / 0.984 $\rightarrow$ \textbf{0.984 / 0.989} & \textbf{0.971} / 0.971 $\rightarrow$ 0.960 / \textbf{0.984} & \textbf{0.934} / 0.809 $\rightarrow$ 0.908 / \textbf{0.847} \\
            Hazelnut    & \textbf{0.989} / 0.954 $\rightarrow$ 0.975 / \textbf{0.964} & \textbf{0.903} / 0.946 $\rightarrow$ 0.901 / \textbf{0.963} & 0.729 / \textbf{0.958} $\rightarrow$ \textbf{0.895} / 0.950 \\
            Leather & \textbf{1.000} / 0.989 $\rightarrow$ \textbf{1.000 / 0.996} & \textbf{1.000} / 0.978 $\rightarrow$ \textbf{1.000 / 0.993} & 0.816 / 0.823 $\rightarrow$ \textbf{0.998 / 0.956} \\
            Metal nut   & 0.817 / 0.804 $\rightarrow$ \textbf{0.851 / 0.893} & 0.765 / 0.915 $\rightarrow$ \textbf{0.767 / 0.939} & 0.918 / 0.545 $\rightarrow$ \textbf{0.933 / 0.835} \\
            Pill    & 0.824 / 0.917 $\rightarrow$ \textbf{0.977 / 0.957} & 0.855 / 0.924 $\rightarrow$ \textbf{0.974 / 0.963} & \textbf{0.995 / 0.938} $\rightarrow$ 0.903 / 0.936 \\
            Screw   & \textbf{0.752} / 0.987 $\rightarrow$ 0.751 / \textbf{0.993} & 0.837 / 0.975 $\rightarrow$ \textbf{0.896 / 0.993} & \textbf{0.883 / 0.927} $\rightarrow$ 0.786 / 0.920 \\
            Tile    & 0.942 / 0.878 $\rightarrow$ \textbf{0.999 / 0.977} & 0.937 / 0.875 $\rightarrow$ \textbf{0.999 / 0.980} & 0.934 / 0.840 $\rightarrow$ \textbf{0.948 / 0.867} \\
            Toothbrush  & 0.992 / 0.979 $\rightarrow$ \textbf{1.000 / 0.991} & 0.964 / 0.964 $\rightarrow$ \textbf{1.000 / 0.986} & 0.770 / 0.883 $\rightarrow$ \textbf{0.919 / 0.957} \\
            Transistor  & \textbf{0.857} / 0.685 $\rightarrow$ 0.840 / \textbf{0.779} & 0.876 / 0.764 $\rightarrow$ \textbf{0.891 / 0.839} & 0.821 / \textbf{0.950} $\rightarrow$ \textbf{0.915} / 0.884 \\
            Wood    & 0.943 / \textbf{0.917} $\rightarrow$ \textbf{0.977} / 0.817 & \textbf{0.955} / 0.900 $\rightarrow$ 0.943 / \textbf{0.904} & 0.875 / \textbf{0.856} $\rightarrow$ \textbf{0.887} / 0.715 \\
            Zipper  & 0.939 / 0.965 $\rightarrow$ \textbf{0.993 / 0.989} & \textbf{0.947} / 0.929 $\rightarrow$ \textbf{0.947 / 0.982} & \textbf{0.970 / 0.967} $\rightarrow$ 0.902 / 0.934 \\
        \hline
        \hline
            Average & 0.900 / 0.923 $\rightarrow$ \textbf{0.935 / 0.946} & 0.916 / 0.926 $\rightarrow$ \textbf{0.941 / 0.959} & 0.887 / 0.860 $\rightarrow$ \textbf{0.902 / 0.905} \\
        \hline
    \end{tabular}
    \renewcommand{\arraystretch}{1}
    }
    \label{table:perform_single}
\end{table}

\subsection{Experiments with other masking methods}
We aim to demonstrate the scalability of FADeR with other single deterministic masking methods by this experiment.
We adopt two defect segmentation models DSR~\cite{DSR_ZavrtanikK_ECCV22} and MSFlow~\cite{MSFlow_Zhou_TNNLS24} and defect token prediction model AMI-Net~\cite{AMINet_Luo_TASC24}.
DSR~\cite{DSR_ZavrtanikK_ECCV22} and MSFlow~\cite{MSFlow_Zhou_TNNLS24} are trained to perform self-supervised defect segmentation based on synthetic defective samples.
AMI-Net~\cite{AMINet_Luo_TASC24} is trained to mask outlier tokens by leveraging token embedding and clustering techniques.

Even if a model trained for defect segmentation is used for masking, the incomplete mask issue will still appear.
For example, their defect segmentation performance is slightly lower than AUROC of 1.0 which means defective regions are still partially missed.
The detail reasons for incomplete mask issues for each model are as follows:
1) The mask of DSR~\cite{DSR_ZavrtanikK_ECCV22} and AMI-Net~\cite{AMINet_Luo_TASC24} tightly fits or smaller than the defect regions.
2) The mask boundaries of MSFlow~\cite{MSFlow_Zhou_TNNLS24} become ambiguous due to multi-resolution fusion.

Experimental results are summarized in Table~\ref{table:perform_single}. 
Before applying FADeR (left side of each `$\rightarrow$'), three models show degraded UAD performances compared to DINO-ViT~\cite{DINO_Caron_ICCV21}.
However, we confirm that both the UAD performance and the defect localization are improved by applying FADeR, shown in each right side of the `$\rightarrow$'.
We cloud easily achieve a performance gain of 1.7\% to 3.8\%  without any change of the pre-trained U-Net.
Thus, we conclude that FADeR effectively resolves the incomplete mask issue regardless of the changes in the masking method.

\begin{table}[t]
    \centering
    \scriptsize
    \caption{Result of UAD on VisA dataset~\cite{VisA_Zou_ECCV22}. The compared models in this experiment show the twice scale of U-Net. Those consist of one NN for normal pattern reconstruction and another NN for defective segmentation.}
    \setlength\tabcolsep{4pt}
    \resizebox{\columnwidth}{!}{%
        \begin{tabular}{l || *{3}{P{1.35cm}} | *{1}{P{1.35cm}}}
            \hline
                \multirow{3}{*}{Model} & \multicolumn{3}{c|}{2-stage} & 1-stage \\
                \cline{2-5}
                & DRAEM & JNLD & OmniAL & FADeR$_{\textit{EAR}}$ \\
                & \cite{Draem_Vitzan_ICCV21} & \cite{JNLD_Zhao_ICME22} & \cite{OmniAL_Zhao_CVPR23} & (ours) \\ 
            \hline
            \hline
                Candle & 0.823 & 0.891 & 0.851 & \textbf{0.927} \\
                Capsules & 0.773 & 0.891 & 0.879 & \textbf{0.919} \\
                Cashew & 0.942 & 0.960 & \textbf{0.971} & 0.967 \\
                Chewing gum & 0.934 & \textbf{0.985} & 0.949 & 0.926 \\
                Fryum & \textbf{1.000} & 0.932 & 0.970 & 0.968 \\
                Macaroni1 & 0.703 & 0.943 & 0.969 & \textbf{0.991} \\
                Macaroni2 & 0.713 & 0.865 & 0.899 & \textbf{0.965} \\
                PCB1 & 0.713 & 0.820 & 0.966 & \textbf{0.967} \\
                PCB2 & 0.897 & 0.963 & \textbf{0.994} & 0.987 \\
                PCB3 & 0.731 & \textbf{0.969} & \textbf{0.969} & 0.974 \\
                PCB4 & 0.913 & 0.948 & 0.974 & \textbf{0.999} \\
                Pipe fryum & 0.941 & \textbf{0.960} & 0.914 & 0.933 \\
            \hline
            \hline
                Average & 0.841 & 0.930 & 0.942 & \textbf{0.960} \\
            \hline
        \end{tabular}
    }
    \label{table:perform_visa}
\end{table}

\subsection{Experiments on another industrial dataset}
To extend the effectiveness of our method beyond the MVTec AD dataset~\cite{MVTec_Bergmann_CVPR19}, we conduct additional evaluations on the VisA dataset~\cite{VisA_Zou_ECCV22}. 
In studies on the VisA dataset~\cite{VisA_Zou_ECCV22}, there are very few cases where a single small NN is used.
Therefore, we compare the performance with a model twice the scale of U-Net~\cite{Draem_Vitzan_ICCV21,JNLD_Zhao_ICME22,OmniAL_Zhao_CVPR23} which is widely used structures in recent studies. 
Those models commonly feature a 2-stage NN architecture: stage 1 performs reconstruction-by-inpainting, and stage 2 conducts defect segmentation. 
The measured AUROCs are summarized in Table~\ref{table:perform_visa}.
Our method achieves the best AUROC of 0.960 as shown in the last column of Table~\ref{table:perform_visa}.

\section{Conclusion}
This study focuses on resolving the incomplete masking issue of single masking in the reconstruction-by-inpainting-based UAD approach.
We propose a simple but powerful NN defective feature attenuation method, FADeR.
FADeR is designed with two-layer MLP to cut out unintended defective information passing caused by an incomplete mask issue.
There is no label to train NN for defective feature attenuation but we also effectively overcome this by exploiting an active learning strategy.

Experimental results on the MVTec AD dataset demonstrated that FADeR outperforms other similar-scale NNs. 
Also, its scalability and effectiveness are further confirmed through integration with various masking methods and industrial datasets.
This suggests the practical feasibility and versatility of FADeR in industrial anomaly detection scenarios.
In conclusion, FADeR offers a reasonable solution by effectively addressing the issues of existing techniques, ensuring better detection accuracy and computational efficiency suitable for edge computing environments.

{\small
\bibliographystyle{ieee_fullname}
\bibliography{egbib}
}

\end{document}